\documentclass[sigconf]{acmart}

\usepackage{booktabs} 
\usepackage{amsmath}
\usepackage{enumitem}
\usepackage{algorithm}
\usepackage{algpseudocode}
\usepackage{subcaption}

\renewcommand\footnotetextcopyrightpermission[1]{} 
\begin{document}

\title{Meta Learning Black-Box Population-Based Optimizers}

\author{Hugo Siqueira Gomes\textsuperscript{1}, Benjamin Léger\textsuperscript{1}, Christian Gagné\textsuperscript{1,2}}
\email{{hugo.siqueira-gomes.1, benjamin.leger.1}@ulaval.ca, christian.gagne@gel.ulaval.ca}
\affiliation{\institution{\textsuperscript{1}Institute Intelligence and Data (IID) - Université Laval, and \textsuperscript{2}Canada CIFAR AI Chair, Mila}}

\renewcommand{\shortauthors}{Gomes et al.}

\begin{abstract}
The no free lunch theorem states that no model is better suited to every problem. A question that arises from this is how to design methods that propose optimizers tailored to specific problems achieving state-of-the-art performance. This paper addresses this issue by proposing the use of meta-learning to infer population-based black-box optimizers that can automatically adapt to specific classes of problems. We suggest a general modeling of population-based algorithms that result in Learning-to-Optimize POMDP (LTO-POMDP), a meta-learning framework based on a specific partially observable Markov decision process (POMDP). From that framework’s formulation, we propose to parameterize the algorithm using deep recurrent neural networks and use a meta-loss function based on stochastic algorithms’ performance to train efficient data-driven optimizers over several related optimization tasks. The learned optimizers’ performance based on this implementation is assessed on various black-box optimization tasks and hyperparameter tuning of machine learning models. Our results revealed that the meta-loss function encourages a learned algorithm to alter its search behavior so that it can easily fit into a new context. Thus, it allows better generalization and higher sample efficiency than state-of-the-art generic optimization algorithms, such as the Covariance matrix adaptation evolution strategy (CMA-ES).
\end{abstract}

\maketitle

\section{Introduction}

\setlength{\textfloatsep}{0.3cm}
\setlength{\abovecaptionskip}{0.3cm}
\setlength{\fboxsep}{0.3pt}
\newcommand\figureonewidth{0.6in}
\newcommand\figureoneheight{0.4in}
\begin{figure}
    \begin{center}
    \setlength{\tabcolsep}{1pt}
	\begin{tabular}{lccccc}
	    \cmidrule{2-6} 
	    \multicolumn{6}{c}{\textbf{Hyperparameter Tuning for SVM}} \\ \cmidrule{2-6}
	      & \tiny{\textbf{GENERATION 1}} & \tiny{\textbf{GENERATION 2}} & \tiny{\textbf{GENERATION 3}} & \tiny{\textbf{GENERATION 15}} & \tiny{\textbf{GENERATION 20}}
		\\ \rotatebox[origin=c]{90}{\textbf{LPBO}} 
		& \raisebox{-.4\height}{\fbox{\includegraphics[width = \figureonewidth, height=\figureoneheight]{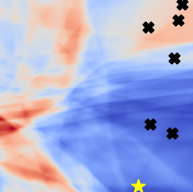}}}
		& \raisebox{-.4\height}{\fbox{\includegraphics[width = \figureonewidth, height=\figureoneheight]{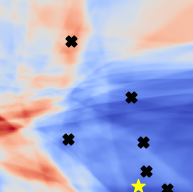}}}
		& \raisebox{-.4\height}{\fbox{\includegraphics[width = \figureonewidth, height=\figureoneheight]{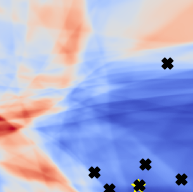}}}
		& \raisebox{-.4\height}{\fbox{\includegraphics[width = \figureonewidth, height=\figureoneheight]{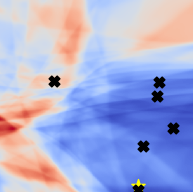}}}
		& \raisebox{-.4\height}{\fbox{\includegraphics[width = \figureonewidth, height=\figureoneheight]{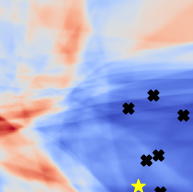}}}
		\\ & \tiny{\textbf{GENERATION 1}} & \tiny{\textbf{GENERATION 5}} & \tiny{\textbf{GENERATION 7}} & \tiny{\textbf{GENERATION 15}} & \tiny{\textbf{GENERATION 20}}
		\\ \rotatebox[origin=c]{90}{\textbf{CMA-ES}} 
		& \raisebox{-.4\height}{\fbox{\includegraphics[width = \figureonewidth, height=\figureoneheight]{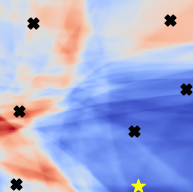}}}
		& \raisebox{-.4\height}{\fbox{\includegraphics[width = \figureonewidth, height=\figureoneheight]{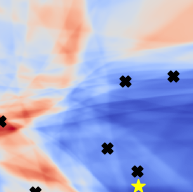}}}
		& \raisebox{-.4\height}{\fbox{\includegraphics[width = \figureonewidth, height=\figureoneheight]{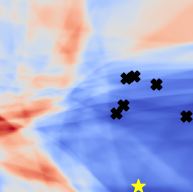}}}
		& \raisebox{-.4\height}{\fbox{\includegraphics[width = \figureonewidth, height=\figureoneheight]{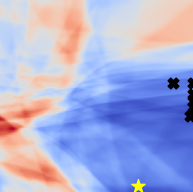}}}
        & \raisebox{-.4\height}{\fbox{\includegraphics[width = \figureonewidth, height=\figureoneheight]{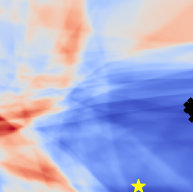}}}
		\\ & \tiny{\textbf{GENERATION 1}} & \tiny{\textbf{GENERATION 10}} & \tiny{\textbf{GENERATION 30}} & \tiny{\textbf{GENERATION 50}} & \tiny{\textbf{GENERATION 298}}
		\\ \rotatebox[origin=c]{90}{\textbf{RS}} 
		& \raisebox{-.4\height}{\fbox{\includegraphics[width = \figureonewidth, height=\figureoneheight]{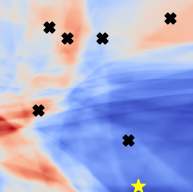}}}
		& \raisebox{-.4\height}{\fbox{\includegraphics[width = \figureonewidth, height=\figureoneheight]{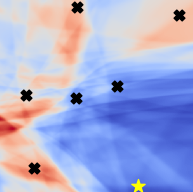}}}
		& \raisebox{-.4\height}{\fbox{\includegraphics[width = \figureonewidth, height=\figureoneheight]{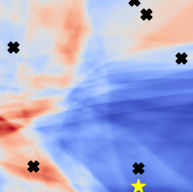}}}
		& \raisebox{-.4\height}{\fbox{\includegraphics[width = \figureonewidth, height=\figureoneheight]{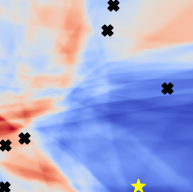}}}
        & \raisebox{-.4\height}{\fbox{\includegraphics[width = \figureonewidth, height=\figureoneheight]{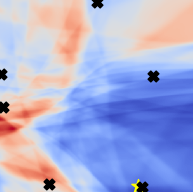}}}
		\\ \\ [-0.5em] \cmidrule{2-6} \multicolumn{6}{c}{\textbf{Training Optimization Tasks}} \\ \cmidrule{2-6} \\ [-1em]
		& {\fbox{\includegraphics[width = \figureonewidth, height=\figureoneheight]{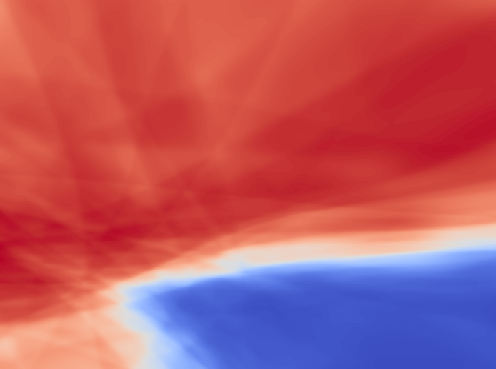}}}
		& {\fbox{\includegraphics[width = \figureonewidth, height=\figureoneheight]{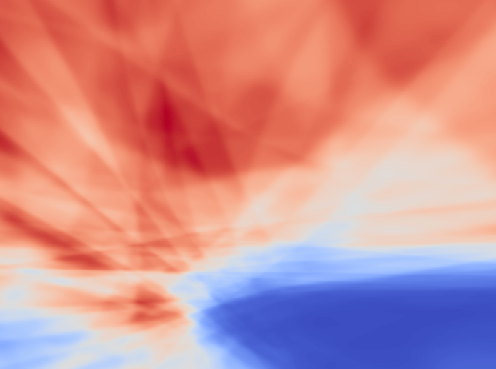}}}
		& {\fbox{\includegraphics[width = \figureonewidth, height=\figureoneheight]{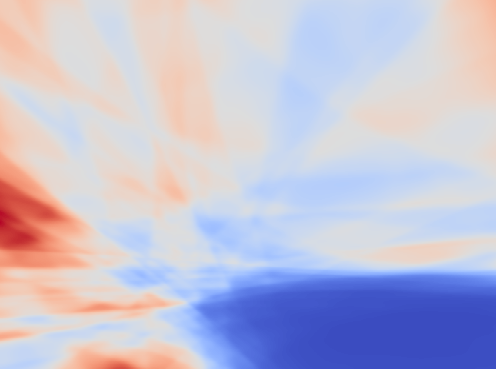}}}
		& {\fbox{\includegraphics[width = \figureonewidth, height=\figureoneheight]{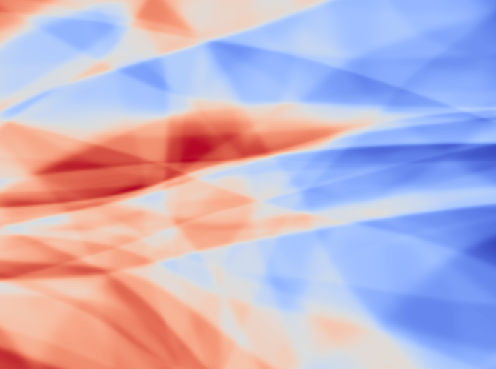}}}
        & {\fbox{\includegraphics[width = \figureonewidth, height=\figureoneheight]{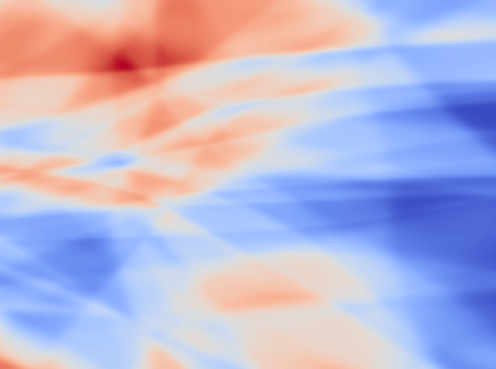}}}
	\end{tabular}
	\caption{An example run of a learned optimizer. Most algorithms may fail to locate the global optimum (yellow star) due to the function's deceptive nature. The learned population-based optimizer (LPBO) solution learns an inductive bias over related training tasks (bottom). It enables a better initialization of the population (black cross, generation 1), a faster adaptation of the search behavior to exploit the function (generation 5), and encourages the maintaining of diversity in the population to reach other regions efficiently without the use of external mechanisms (e.g., restart strategies or diversity preservation)}
	\label{fig:intro} 
    \end{center}
\end{figure}

Solving Black-Box Optimization (BBO) problems is usually addressed in two main ways: either using a generic black-box algorithm to solve a problem or developing a custom optimizer that includes specific elements related to the nature of the problem. Generic BBO algorithms are appealing since they can be applied to a given problem with relatively little adaptation efforts. Then, the search rules may be flexible enough to overcome the given problem's complexity and constraints. However, their search can be highly inefficient, requiring a high number of solutions to be evaluated, and their performance may be relatively limited -- the notion of ``good enough'' solutions depending on the task to be solved. As for custom optimization methods, they might be effective in many cases, but they may rely on expert knowledge and are likely to require laborious work to be validated, often with specific implementation due to the problem's constraints. There may be a high cost of effort and time to invest in proposing and validating new search rules that fit the nature of the problem.

In this paper, we follow an alternative path in-between these two avenues. We propose to automatically learn custom algorithms over several instances of related optimization tasks in a data-driven fashion. This way, we take the best of both worlds by achieving a method that requires little adaption efforts to the problem complexities while creating optimizers that are tailored to the specific problems at hand, with both strong performance and good sample efficiency. This strategy is tricking the No Free Lunch theorem \citep{no_free_lunch_theorem} to some extent -- which states that no model is superior to others for every possible problem -- since we meta-optimize an algorithm in the space of optimizers in order to find search rules that are particularly fit for a specific class of problems. This is known as \emph{learning to optimize} \citep{li_learning_2017}.

Learning to optimize has been addressed with reinforcement learning \cite{chen_learning_2017} and gradient-based approaches \cite{wang_learning_2017, andrychowicz_learning_2016}. Here, we propose to increase the adaptability of the learned optimizer by using a common structure of population-based optimizers to fit BBO scenarios. We introduce Learning-to-Optimize POMDP (LTO-POMDP), a meta-learning framework that automatically learns a population-based algorithm over a given distribution of optimization problems. The meta-learning component of our approach aims to find an optimizer model over a set of related tasks with the right inductive bias to quickly adapt its search rules to a new unseen optimization task from the same type of problems. 

The purpose of this paper is to highlight meta-learning popula\-tion-based algorithms as a compelling option to the black-box optimization toolbox over three main contributions:
\begin{itemize}
    \item A meta-learning formulation based on partially observable models that enables learning po\-pu\-la\-ti\-on-based optimizers from scratch given a distribution of optimization tasks;
    \item A general methodology for training and assessing performances of learned black-box algorithms;
    \item A specific implementation of a novel policy and meta-loss function for learned black-box algorithms, with conclusive results on several BBO problems (i.e., optimization benchmarks and hyperparameter optimization of machine learning models).
\end{itemize}

\section{Related Work}
\label{sec:2.0}

This work is connected to population-based metaheuristics, meta-learning, and, more specifically, meta-reinforcement learning.

\textbf{Population-based metaheuristics} \cite{talbi_metaheuristics_2009} encompass a range of algorithms that perform relatively well on various types of optimization tasks. They usually do not require any property of the underlying objective function and rely on relatively permissive assumptions of the search space. Although they are often used in classical BBO scenarios, recent works have shown their effectiveness for training in high dimension \cite{such_deep_2017}, reinforcement learning models \cite{salimans_evolution_2017} and neural architecture search \cite{lu_nsga-net_2019}. They tend to be a competitive choice in terms of performance, especially when the number of time steps in an optimization episode is long \cite{metz_tasks_2020}. Besides, they are implicitly parallel, which is suitable for execution on massive clusters and allows a joint exploration of the search space, offering greater resilience over the highly non-convex multi-modal nature of many real-life problems. In particular, the recent trend of quality-diversity optimization algorithms \cite{costa_exploring_2020, mouret_quality_2020} indicates the importance of different individuals during the optimization process. In this work, we are using population-based algorithms given their capacity to adapt to a given context, which is required to generate effective and robust general optimizers. We use them as a high-level template of the search space of learned optimizers.

\textbf{Meta Learning or Learning to learn} is a promising approach that aims to train a machine learning model on various learning tasks to achieve considerable generalization over new unseen tasks \cite{ravi_optimization_2017, finn_model-agnostic_2017, snell_prototypical_2017, vinyals_matching_2016}. There are variants of the problem that have been proposed for similar contexts such as \emph{multitask learning} \cite{caruana_multitask_1997} or \emph{lifelong learning} \cite{beaulieu_learning_2020}. In this work, we are interested in meta-learning in the context of \emph{learnable optimizers} (i.e., learning to optimize) \cite{li_learning_2017, cao_learning_2019}. The learning scenario requires training an optimizer over a set of related functions to achieve a general search that is robust and fast to optimize new unseen functions. Metz et al. \cite{metz_tasks_2020} address this problem as a bilevel optimization problem, considering a base (or inner) loop that uses an optimizer to solve a particular function and the meta (or outer) loop that updates the learned optimizer parameters. Another approach was proposed in Wang et al. \cite{wang_learning_2017} to train a recurrent neural network to represent its own reinforcement learning procedure. Other works have replaced variants of gradient descent for learned gradient-based optimizers \cite{andrychowicz_learning_2016, chen_learning_2017}. Although machine learning is often centered on gradient-based optimizers, this work does not require any additional information about the training or testing functions. In other words, the learned optimizer of this paper can be applied to black-box optimization tasks (i.e., it does not require gradient information) with competitive performance.

\textbf{Meta-Reinforcement Learning} is a research area where meta-learning is applied to reinforcement learning (RL). Since RL needs to scale on increasingly complex tasks with costs evaluated in the real-world, sample efficiency has become a critical issue. Meta-RL then addresses the problem of increasing the generality of RL algorithms to new tasks or environments not encountered by the model during training \cite{duan_rl2_2016}. The problem setup is similar to RL, but the agent is trained on a distribution of Markov decision processes (MDPs), and the policy is a model with memory where it receives the last reward and the last action at each state. MAML \cite{rajeswaran_meta-learning_2019} and Reptile \cite{nichol_first-order_2018} provide an example of how to approach this kind of problem. Recent works propose to update the model parameters with respect to a meta-loss function to achieve good generalization performance on unseen tasks. Schulman et al. \cite{schulman_high-dimensional_2016} propose learning the loss function directly to achieve higher returns at test time. Humplik et al. \cite{humplik2019meta} formally define the problem using a partially observable Markov decision process (POMDP) that allows an agent to train a belief module to predict the task information in a supervised manner. In this work, we also formalize our learning problem using a POMDP in a similar direction. However, instead of training RL algorithms in the POMDP for a distribution of MDPs, we train general-purpose optimizers in our POMDP for a distribution of related problems of interest.

\section{Revisiting Population-Based Search}
\label{sec:Revisiting-Population-Based-Search}

In this section we introduce the underlying structure of population-based algorithms that will serve as the basis of our meta-learning framework. We consider a BBO problem:
\begin{equation}
\label{Eq:bbo}
x^* = \mathop{\arg\min}_{x \in \mathbb{R}^d} f(x),
\end{equation}
where the objective function $f: \mathcal{X} \rightarrow \mathbb{R}$ can be evaluated at each point $x \in \mathcal{X}$ and no other information about $f$ is available (e.g., no analytical form, gradient). In this problem, an iterative optimizer outputs a time-ordered sequence of evaluations:
\begin{equation}
f(x_1), f(x_2), \ldots, f(x_H),
\label{Eq:trajectory_1}
\end{equation}
where $H$ is the maximum number of evaluation steps done by the optimizer for a given problem $f$ and each point $x_1, x_2, \ldots, x_H$ is generated by a decision-making process of the optimizer. Two main approaches to search are usually considered: single-solution sequential search or population-based search \cite{talbi_metaheuristics_2009}. To simplify, we assume that the difference between them is how the optimizer generates an evaluation path. In the case of population-based search, the target function is typically evaluated using a batch of $\lambda$ points whereas single-solution based algorithms manipulate and transform a single solution during the search ($\lambda=1$). Indeed, many proposed search mechanisms are similar between them, and certain hybrid approaches may also be considered. Let $G$ be the total number of populations evaluated (i.e., number of generations), then we can rewrite Eq.~\ref{Eq:trajectory_1} for both cases as:
\begin{equation*}
{\{f(x_1), \ldots, f(x_{\lambda})\}}_{0}, {\{f(x_1), \ldots, f(x_{\lambda})\}}_{1}, \ldots, {\{f(x_1),\ldots,f(x_{\lambda})\}}_{G}.
\end{equation*}
Let $P^g = \{P^g_X,P^g_Y\}$ be the population at generation $g$ where $P^g_X = \{ x_{0}, \ldots, x_{\lambda}\}_g$ is the set of points on the search space and $P^g_Y = \{ f(x_{0}), \ldots, f(x_{\lambda})\}_g$ is the set of function evaluations at those points. Therefore, we can describe a general \emph{optimization rollout} as a sequence of search points and evaluations over time:
\begin{equation}
\label{Eq:trajectory_3}
P_X^{0}, P_Y^{0}, P_X^{1}, P_Y^{1}, \ldots, P_X^{G}, P_Y^{G}.
\end{equation}

\setlength{\textfloatsep}{0.5cm}
\begin{algorithm}[t]
\caption{General Template of Population-Based Algorithms}\label{Alg:population-based-template}
\begin{algorithmic}[1]
\Procedure{optimize}{$f$}
\State $P_X^{0} \leftarrow \text{generate initial population}$ 
\State $P_Y^{0} = f(P_X^0)$
\For{g = 1, 2, \ldots, G}
    \State {$P_X^{g} \sim \pi(P_X | \{ P_X^i, P_Y^i\}_{i=0}^{g-1})$}
    \State {$P_{Y}^g = f(P_{X}^g)$}
\EndFor
\EndProcedure
\label{pop-search}
\end{algorithmic}
\end{algorithm}

Specifically, a population-based search approach is described in Algo.~\ref{Alg:population-based-template}. Population-based optimization algorithms follow this update formula to the population at each time step:
\begin{equation}
P_X^{g} \sim \pi(P_X | \{ P_X^i, P_Y^i\}_{i=0}^{g-1}),
\label{Eq:pi}
\end{equation}
where $\pi$ defines a distribution of probability over different search points $P_X$; and $\{ P_X^i, P_Y^i\}_{i=0}^{g-1}$ is the \emph{optimization rollout} at generation $g-1$ (i.e., all search points and evaluations done so far). The optimization starts with the first set of points $P_X^0$ (known as \emph{initial population}) (line 2). Then, the function is evaluated at those points (line 3). The optimizer receives the respective evaluations and makes the decision process to generate the next batch of points to be evaluated (line 5). The procedure continues until a \emph{stopping criteria} (e.g., maximum budget) is reached (line 4). 

In practice, optimizers can be specified by hand-designed heuristics and search mechanisms represented by $\pi$. Thus, $\pi$ defines the decision process to generate new points to be evaluated. For example, Evolutionary Algorithms (EAs) \cite{kenneth_unified_2016, xue_survey_2016, xue_gabased_2020} often implement replacement, selection, mutation, and crossover operators to generate the next search points $P^g_X$ (the \emph{offspring}). In the case of Estimation of Distribution Algorithms (EDAs) \cite{zhang_convergence_2004, hauschild_introduction_2011}, a probabilistic distribution may be used to represent the current population to reproduce a new \emph{offspring}. In Particle Swarm Optimization (PSO) \cite{wang_particle_2018}, each particle's previous position and velocity are used to update $P^g_x$ as the new position of each particle. An in-depth explanation and theoretical analysis of population-based search mechanisms can be found in \cite{talbi_metaheuristics_2009}.

In this work, instead of proposing new handcrafted search mechanisms for specific problems of interest, we are looking to learn the optimizer automatically. The goal is to find a parameterization $\theta$ for $\pi(\cdot)$ (from now on $\pi_{\theta}$) by improving it based on the resulting performance on a set of training optimization tasks to learn general search rules for related problems. Regardless of whether one decides to use handcrafted heuristics or learn the search rules, most if not all population-based algorithms fit into this perspective: $\pi_\theta$ represents the optimizer that acts as a decision-making solver.

The interpretation of $\pi_\theta$ can be extended to consider other parts of population-based algorithms (e.g., initial population strategy, stopping criteria, search space representation). In other words, the representation of $\pi_\theta$ will determine which part of a population-based algorithm is learnable. In addition to the decision-making process (Eq.~\ref{Eq:pi}), we also chose to learn the \emph{initial population strategy}. However, we use a fixed number of steps as \emph{stopping criteria} and a continuous space as the \emph{representation of the search space} in all of our experiments described in Sec. \ref{sec:5}.
\section{Learning Population-Based Optimizers}

In the following, the Learning-to-Optimize POMDP (LTO-POMDP) meta-learning framework is presented.  We first formalize the proposed framework through the POMDP paradigm \cite{krishnamurthy2016partially} (Sec.~\ref{sec:4.1}). Then, the rest of this section describes the meta-objective function and the training data of our meta-learning framework: how to measure the performance of learned optimizers (Sec.~\ref{sec:4.2}) and how to define task distributions (Sec.~\ref{sec:4.3}). 

\begin{center}
\begin{algorithm}[tb]
\caption{A single run in Learning-to-Optimize POMDP}\label{Alg:Learning-to-Optimize-POMDP}
\begin{algorithmic}[1]
\State At time $t = 0$:
\State \quad current task $w_k \sim p(w)$ is sampled
\State \quad initial state set to $s_0 = \{\emptyset, w_k\}$
\State \quad initial observation set to $o_0 = \{\emptyset\}$ 
\For{t = 0,1,2,\ldots,T}
    \State policy takes an action $a_t$ according to current
    
    observation $o_t$ and available information $\mathcal{I}_t$:
    \begin{align*}
        a_t &\sim \pi_{\theta}(a | o_t, \mathcal{I}_t) \in \mathcal{A} \\
        \text{with} \ \mathcal{I}_t &= \{\emptyset, a_0, r_0, \ldots, o_{t-1}, a_{t-1}, r_{t-1}\}
    \end{align*}
    \State policy receives the reward $r_t = r(s_t, a_t)$ for choosing $a_t$ 
    \State update state based on the current task $w_k$
    \begin{align*}
        s_{t+1} = \{w_k(a_t), w_k\}
    \end{align*}
    \State generate observation deterministically from the state
    \begin{align*}
        o_{t+1} = \{w_k(a_t)\}
    \end{align*}
    \State policy updates available information $\mathcal{I}_{t+1}$ 
    \begin{align*}
        \mathcal{I}_{t+1} = I_t \cup \{ a_t, r_t, o_{t+1} \}
    \end{align*}
\EndFor
\State Final policy reward $r_T = r(s_T, a_T)$ received
\end{algorithmic}
\end{algorithm}
\end{center}

\subsection{Learning-To-Optimize POMDP}
\label{sec:4.1}

Learning to optimize can be viewed as a bilevel optimization problem \cite{metz_tasks_2020}. A base (inner) loop is defined by the optimization rollouts of a parametrized optimizer on a single task and a meta (or outer) loop that corresponds to successive updates of its parameters based on the feedback provided by a meta-objective function.

\begin{figure}
    \includegraphics[width=\linewidth]{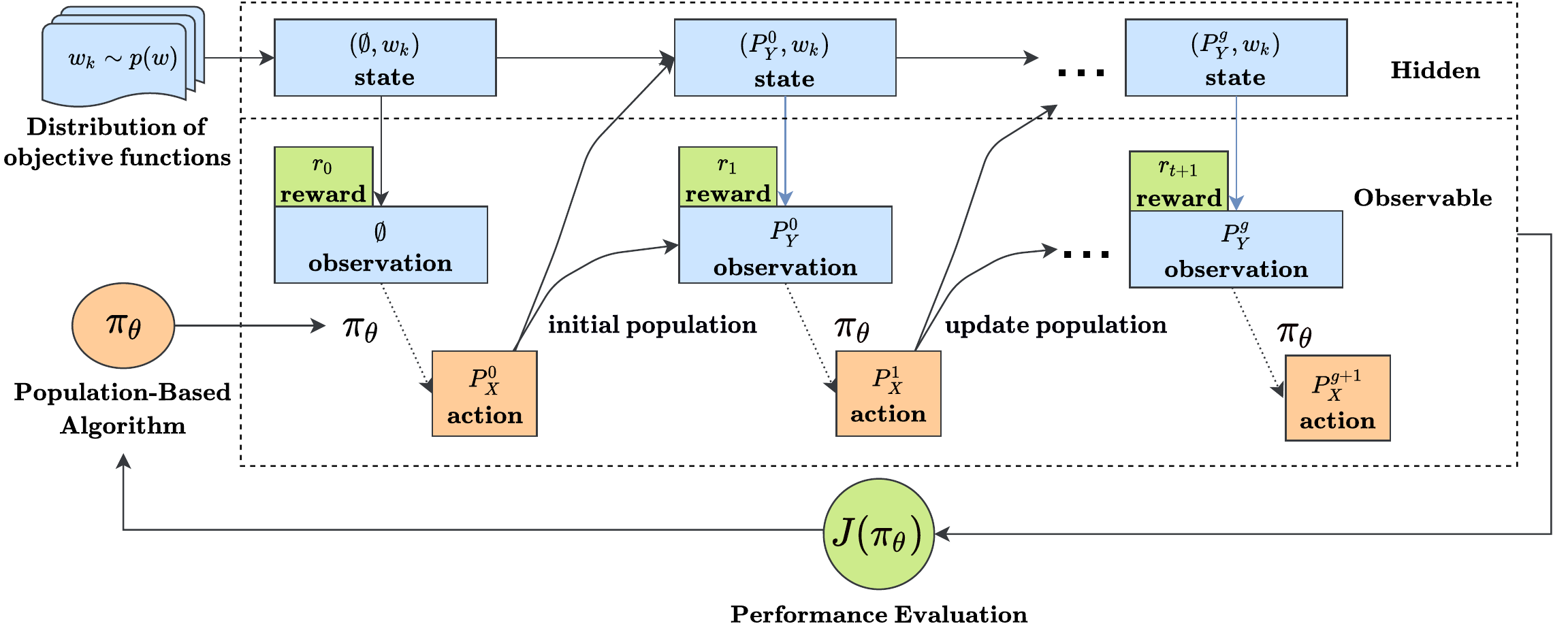}
    \caption[Dynamics of the Learning-to-Optimize POMDP (LTO-POMDP)]{Overview of the learning-to-optimize POMDP (LTO-POMDP).}
    \label{fig:Learning-To-Optimize-POMDP}
\end{figure}

\textbf{Inner-optimization} The execution of a parameterized po\-pu\-la\-tion-based algorithm $\pi_\theta$ is viewed as the execution of a policy in the LTO-POMDP. Algo.~\ref{Alg:Learning-to-Optimize-POMDP} describes a single run of the inner-loop relying on policy $\pi_\theta$. Let $\mathcal{W}$ be a set of tasks and $p(w)$ be a distribution over these tasks $w \in \mathcal{W}$. At time $t=0$, the run begins by sampling a task $w_k \sim p(w)$ used for the whole episode. The current hidden state $s_t$ is defined as a tuple $\{o_t, w_k\}$ where $o_t$ is the current observation. In other words, this particular POMDP defines a task-fixed episode using the optimization task $w_k$ as the only unobserved state. This task then defines the dynamics of what the optimizer observes (i.e., the fitness of new points) during the optimization rollout. Those dynamics of the LTO-POMDP are depicted in Fig.~\ref{fig:Learning-To-Optimize-POMDP}.

Each component of the LTO-POMDP (i.e., the states, observations, actions and rewards) is based on a standard population-based search (Algo.~1). Specifically, we consider a hidden state $s_t = \{P_Y^{g-1}, w_k\}$ that generates deterministically the observation $o_t = \{P_Y^{g-1}\}$ as the evaluations of the previous search points. The optimizer $\pi_\theta$ provides the current action $a_t \sim \pi_{\theta}(a | o_t, \mathcal{I}_t)$ by selecting the search points to be evaluated according to the available information $\mathcal{I}_t = \{ \emptyset, P_X^0, r_0, P_Y^0, P_X^1, r_1, P_Y^1, \ldots, P_Y^{g-1}, P_X^{g-1}, r_{t-1} \}$. It fits well to the context of black-box optimization, where the algorithm does not have access to any information about the task $w_k$ that it is currently optimizing and it observes only the evaluations of the selected search points. Finally, the reward $r(\cdot)$ corresponds to a measure of the algorithms’ performance, explained in further details in Sec.~\ref{sec:4.2}, and is used at the outer (meta) optimization level.

\begin{figure*}
\begin{center}
    \includegraphics[width=\linewidth]{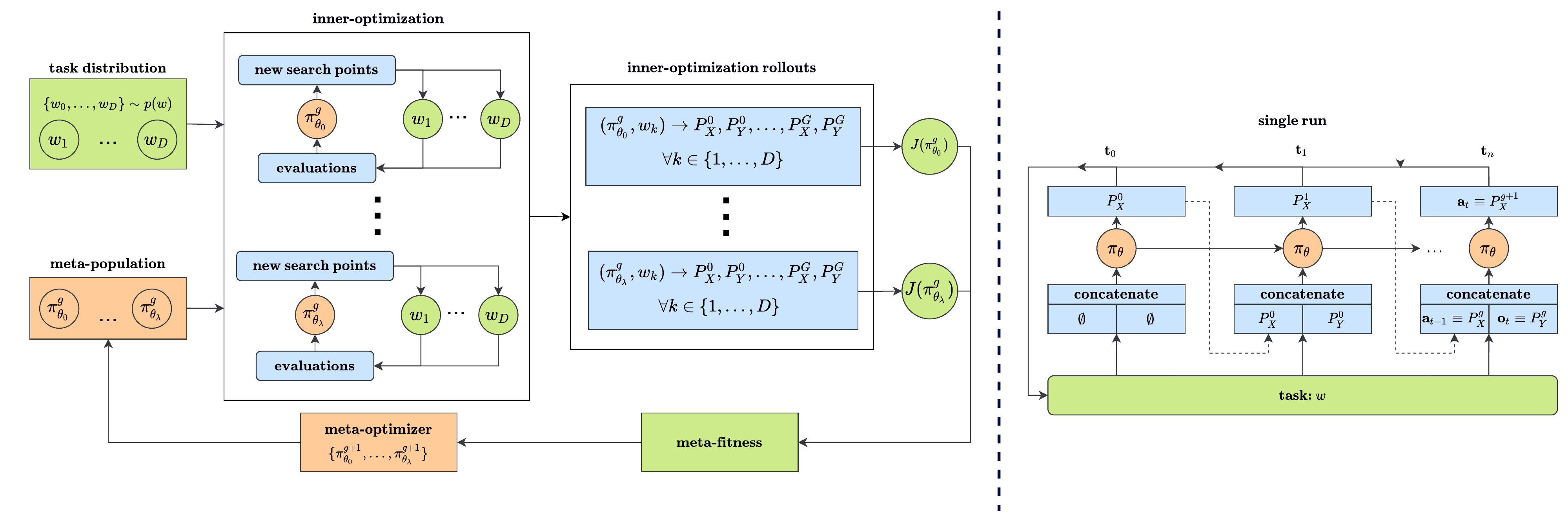}
    \caption{The two-level overview of the system. The outer-optimization loop (\textbf{left}) consists in improving a population of learned optimizers based on each performance $J(\pi_{\theta})$ (green) through successive runs on a set of tasks. The inner-optimization loop (\textbf{right}) uses the learned optimizer $\pi_\theta$ (red) to propose search points through successive interactions with the task (green).}
    \label{fig:OverviewFramework}
\end{center}
\end{figure*}

\textbf{Outer-optimization}
On top of this inner-optimization level, which defines the dynamic of single optimization runs, we describe the \emph{outer-optimization} loop. The learned optimizer $\pi_\theta$ needs to interact with LTO-POMDP to improve its performance over time. Therefore, we need to define a way to evaluate its performance. At this level, parameters $\theta$ of a policy $\pi_{\theta}$ are trained to optimize the overall performance across tasks sampled from $p(w)$. This is done according to a \emph{meta-loss function} that is defined as:
\begin{equation}
\label{Eq:meta-loss}
\min_{\theta}  \displaystyle \mathop{\mathbb{E}}_{w_k \sim p(w)} \left [ J^k(\pi_{\theta}) \right ],
\end{equation}
where $J^k(\pi_{\theta})$ is the performance of the policy $\pi_\theta$ for the task $w_k \sim p(w)$.
The general configuration of the training procedure in the LTO-POMDP is presented in Fig.~\ref{fig:OverviewFramework}. This modeling is relatively flexible and can be adapted in practice according to the optimization tasks distribution (green) and the performance measurement of multiple runs to determine the performance $J(\cdot)$ (blue). In the following subsections, we will describe both elements.

\subsection{Performance Measurement} 
\label{sec:4.2}

In our context, $\pi_\theta$ represents a parameterized (possibly stochastic) learned optimizer. We want to measure its performance and increase it over training time. There are many possible ways to define the performance of a stochastic population-based algorithm. Recent benchmarking platforms \cite{nevergrad, doerr2018iohprofiler, hansen2020coco} have proposed the use of the \textit{expected runtime of the restart algorithm} \cite{auger2005performance}. This serves as the basis of our meta-objective, which can be calculated as follows.

When comparing different optimizers, we should be aware that some optimizers may have a small probability of solving a task but then achieve it quickly, while others may have a higher probability of success but will require more time. Therefore, for a fair evaluation of optimizers, we follow the derivations of equations in Auger and Hansen \cite{auger2005performance} and consider a widely used approach in benchmarking stochastic optimizers: the conceptual \textit{restart algorithm}. We calculate the performance of $\pi_\theta$ as the aggregate of the expected number of function evaluations (FE) to reach a certain function value (\textit{success criterion}) of $w_k$ by simulating independent restarts of $\pi_\theta$ from various runs.

Let $p_s \in (0, 1]$ be the probability of success of $\pi_\theta$ to reach a certain function value of $w_k$ and $\text{FE}_{\text{max}}$ be the maximum number of function evaluations, i.e., the number of evaluations done when $\pi_\theta$ does an unsuccessful run. Then, the performance $J(\pi_\theta, w_k) = \text{FE}(\pi_\theta, w_k) = \text{FE}_{\theta, w_k}$ is a random variable measuring the number of function evaluations until a \emph{success criterion} is met by independent restarts of $\pi_\theta$ on the task $w_k \sim p(w)$:
\begin{equation}
\label{Eq:FE}
\text{FE}_{\theta, w_k} = \sum_{i=1}^{N} \text{FE}_{\text{max}} + \text{FE}^{\text{succ}}_{\theta,w_k},
\end{equation}
where $N$ is the random variable that measures the number of unsuccessful runs of $\pi_\theta$ required to reach once the success criterion and $\text{FE}^{\text{succ}}_{\theta,w_k}$ as the number of evaluations for the successful run of $\pi_\theta$ to reach the criterion.

We now need to look at evaluating the expectation $\mathbb{E}[\text{FE}_{\theta, w_k}]$. First, we write the conditional expectation of Eq.~\ref{Eq:FE} w.r.t $N$:
\begin{equation}
\label{Eq:expected-FE-given-N}
\mathbb{E}[\text{FE}_{\theta, w_k} | N] = N\,\text{FE}_{\text{max}} + \mathbb{E}[\text{FE}^{\text{succ}}_{\theta,w_k}]
\end{equation}
 
Now we use the fact that $N \sim \mathrm{NB}(r = 1, p = 1-p_s)$ follows a negative binomial distribution and its expectation is $\mathbb{E}(N) = \frac{rp}{1-p} = \frac{1(1 - p_s)}{1 - (1 - p_s)} = \frac{1 - p_s}{p_s}$, where $p_s$ is the probability of success of a run. Then, we take the expectation again to write the general equation of $\mathbb{E}[\text{FE}_{\theta, w_k}]$:
\begin{align}
\mathbb{E}[\text{FE}_{\theta, w_k}] &= (\mathbb{E}[N]) \text{FE}_{\text{max}} + \mathbb{E}[\text{FE}^{\text{succ}}_{\theta,w_k}] \\
\label{Eq:expected-FE}
&= \left ( \frac{1 - p_s}{p_s} \right ) \text{FE}_{\text{max}} + \mathbb{E}[\text{FE}^{\text{succ}}_{\theta,w_k}] .
\end{align}

The probability of success $p_s$ and  the expected number of function evaluations for successful runs $\mathbb{E}[\text{FE}^{\text{succ}}_{\theta,w_k}]$ are not available. Therefore, we estimate each of those terms as \cite{suganthan2005problem}: 
\begin{align}
\label{Eq:estimator-p_s}
\hat{p}_s &= \frac{\text{\# successful runs}}{\text{\# runs}},\\ 
\label{Eq:estimator-fesucc}
\widehat{\mathbb{E}[\text{FE}^{\text{succ}}_{\theta,w_k}]} &= \frac{\text{\# total evaluations for successful runs}}{\text{\# successful runs}}.
\end{align}

Finally, using the previous meta-objective (Eq.~\ref{Eq:meta-loss}), the expected number of evaluations (Eq.~\ref{Eq:expected-FE}) and its estimators (Eq.~\ref{Eq:estimator-p_s} and \ref{Eq:estimator-fesucc}), the final meta-objective can be written as:
\begin{align}
\min_{\theta}  \displaystyle \mathop{\mathbb{E}}_{w_k \sim p(w)} & \left [ J(\pi_{\theta}, w_k) \right ] = \min_{\theta}  \displaystyle \mathop{\mathbb{E}}_{w_k \sim p(w)} \left [ \text{FE}_{\theta, w_k} \right ] \nonumber\\
= & \min_{\theta}  \displaystyle \mathop{\mathbb{E}}_{w_k \sim p(w)} \left [ \left ( \frac{1 - \hat{p}_s}{\hat{p}_s} \right ) \text{FE}_{\text{max}} + \widehat{\mathbb{E}[\text{FE}^{\text{succ}}_{\theta,w_k}]} \right ].\label{Eq:meta-objective}
\end{align}

Our experiments (detailed later) demonstrate that this encourages the learned policy to be sample efficient (i.e., solving multiple tasks in a few steps) and robust to premature convergence (i.e., solving more tasks in more steps). Note that other works have used more traditional reward functions \cite{metz_tasks_2020, chen_learning_2017, cao_learning_2019, li_learning_2017}, which can also be used in our proposed model, leading to possibly different behaviors on each class of problems.
\subsection{Task Distribution}
\label{sec:4.3}

The last component of the proposed meta-learning framework relates to the concept of tasks and their distribution. The task distribution in our context can describe any class of optimization problems including hyperparameter optimization \cite{metz2020using}, black-box optimization \cite{hansen2020coco} or reinforcement learning problems \cite{salimans_evolution_2017}. We will consider only continuous optimization problems for the experiments in the remainder of the paper, given their high applicability to many real-world tasks and comparability through benchmarking with other population-based approaches.

Let $f_\nu$ be an objective function and $\nu_i$ the configurations of an instance indexed by $i = 1, 2...$. In order to be consistent with the definition of a task in the meta-learning literature, let $\mathcal{F}$ be the space of objective functions and $\mathcal{V}$ be the space of configurations of objective functions. We consider that a task instance is drawn from a distribution $p(\mathcal{C})$ over a class of optimization problems, such that $w_k \sim p(\mathcal{C})$, where $\mathcal{C} = \mathcal{F} \times \mathcal{V}$.

In practice, we often do not have access to this distribution, and drawing a task may consist of picking a specific (parameterized) instance of a given function family (e.g., a quadratic function, accuracy of SVM) from a set of available functions. The configuration $\nu_i$ can represent coefficients as well as different transformations of the function $f_\nu$. We consider different instances to correspond to variations of the same problem. For example, in synthetic functions, optimal values can be shifted or applied rotation and translation on the search space. In hyperparameter tuning, we can see different instances as different losses that vary based on different datasets or training regimes. The notion of the objective function $f_\nu$ and its configuration $\nu_i$ is thus flexible and defines a problem class. Finally, we represent a task $w$ as a tuple $(f_\nu, \nu_i)$.

\section{Experiments}
\label{sec:5}

In this section, we assess the proposed approach in different scenarios. We begin by presenting the design choices of the policy and the meta-optimizer. The training and evaluation procedure are detailed, and an overview of the problem settings and the optimization baselines are used as a comparison. Finally, we show the empirical results to demonstrate the performance and efficiency of the approach for learning custom optimizers to different related problems. \footnote{The implementation of the framework and all scenarios is available at \url{https://github.com/optimization-toolbox}.}

\subsection{Experimental and Implementation Setup}

\subsubsection{Policy Architecture} 

The trained optimizer uses the same architecture (including hyperparameters) in all experiments. It consists of a recurrent neural network composed of two layers of LSTMs \cite{hochreiter_Long_1997} neurons, with 32 neurons in the hidden layer, followed by a single layer Bayesian Neural Network \cite{bayes_nn} as output. The LSTM layers provide a memory component that enables the modeling of the dependency on all previous populations, which is an architecture choice commonly made in POMDP-based meta-reinforcement tasks \cite{humplik2019meta, duan_rl2_2016}. The Bayesian Neural Network is used to represent stochastic policies because most population-based algorithms have stochastic search mechanisms. The domain search has been unified for all tasks to $[-1,1]$. Therefore, we use ``tanh'' as an activation function for the output. All parameters are initialized using a normal distribution $\theta_0 \sim \mathcal{N}(0,0.5)$.

\begin{figure}
\begin{center}
    \captionsetup{type=figure}
    \includegraphics[width=\linewidth]{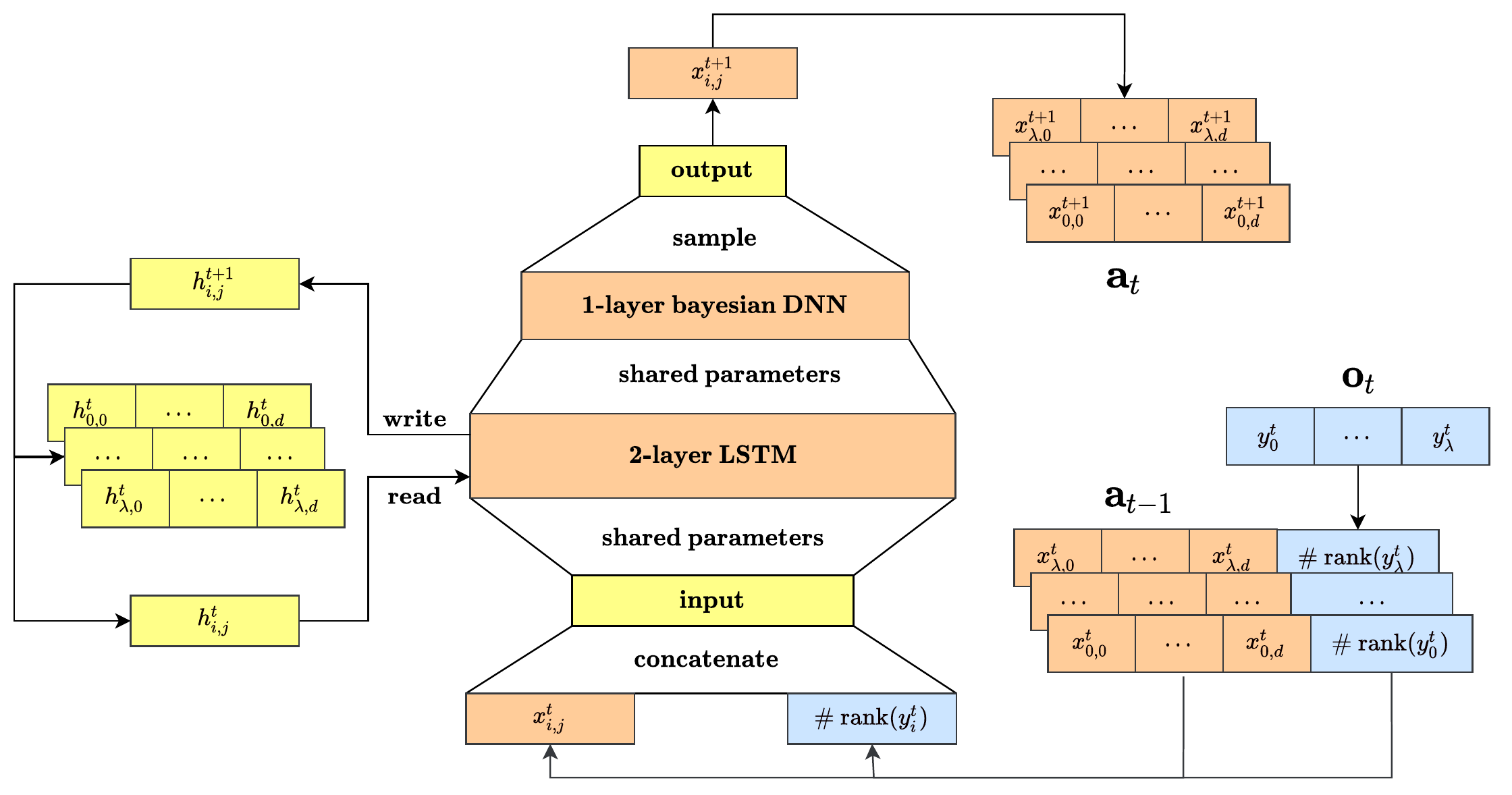}
    \caption{Policy Architecture. The population is updated based on the individual (value, fitness ranking) pair.}
    \label{fig:policy}
\end{center}
\end{figure}

Two additional aspects are taken into consideration in the policy implementation. First, the \emph{scalability} according to the problem dimensionality can be limited. Hence, similarly to recent works on learnable optimizers \cite{andrychowicz_learning_2016, li_learning_2017, cao_learning_2019}, we implement a \emph{coordinate-wise policy}, such that it works independently over the dimensions of the problem. Different memory units (hidden states) are defined for each individual-dimension pair, resulting in more tractable $\lambda \times d$ predictions at each population update -- $\lambda$ being the population size and $d$ the dimensionality of the problem. Another major design option for many optimization algorithms refers to \emph{invariance}, that is, the ability of a method to generalize from a single problem to a class of problems. More specifically, we consider the invariance to monotonically increasing transformations of $f$. For example, the performance of the optimizer in $f$ is the same in $f^3$, $f \times 2\,|f|^{\frac{-5}{9}}$, etc. Therefore, we replace all of the evaluations on $f$ by an adaptive transformation of $f$ to represent how the observed values are relative to other observations in the current step \cite{igo_article}. The implementation uses the ranking of the evaluations instead of their absolute value. These choices allow us to considerably reduce the number of parameters in the model to be optimized since the inputs are only two values and the output a single one (see Fig.~{\ref{fig:policy}}).

\setlength{\textfloatsep}{0.2cm}
\setlength{\abovecaptionskip}{0.2cm}

\subsubsection{Meta-Optimizer}

BBO is a natural choice for optimizing the non-differentiable meta-loss function proposed (Eq.~\ref{Eq:meta-objective}). A slightly modified version of DeepGA \cite{such_deep_2017} is used, which was demonstrated to be efficient in POMDP environments \cite{salimans_evolution_2017}. Specifically, this is a genetic algorithm that maintains a population of parameters $(\theta_1, \hdots, \theta_{\lambda})$ represented by a list of seeds. The seeds define the mutations applied to each individual over generations. Thus, each CPU core can evaluate each policy in the population separately. Instead of sending thousands of thousands of parameters to each other, it can communicate using a list of seeds corresponding to a parameter $\theta_i$. The fitness evaluation is the performance measurement explained in Sec.~\ref{sec:4.2} applied to a batch of functions. We use 512 as the population size, 5 as the number of elite individuals, 20 as the number of parents in each generation and $\sigma_0 = 0.3$. The difference from the original paper is that they use a fixed $\sigma$ in all generations while we use a decaying strategy where we update $\sigma_t = 0.95\times\sigma_{t-1}$ with a minimum $sigma_{\text{min}} = 0.01$ over 200 generations.

\begin{figure*}[!htb]
\centering
 \begin{subfigure}{0.25\textwidth}
   \includegraphics[width=\linewidth]{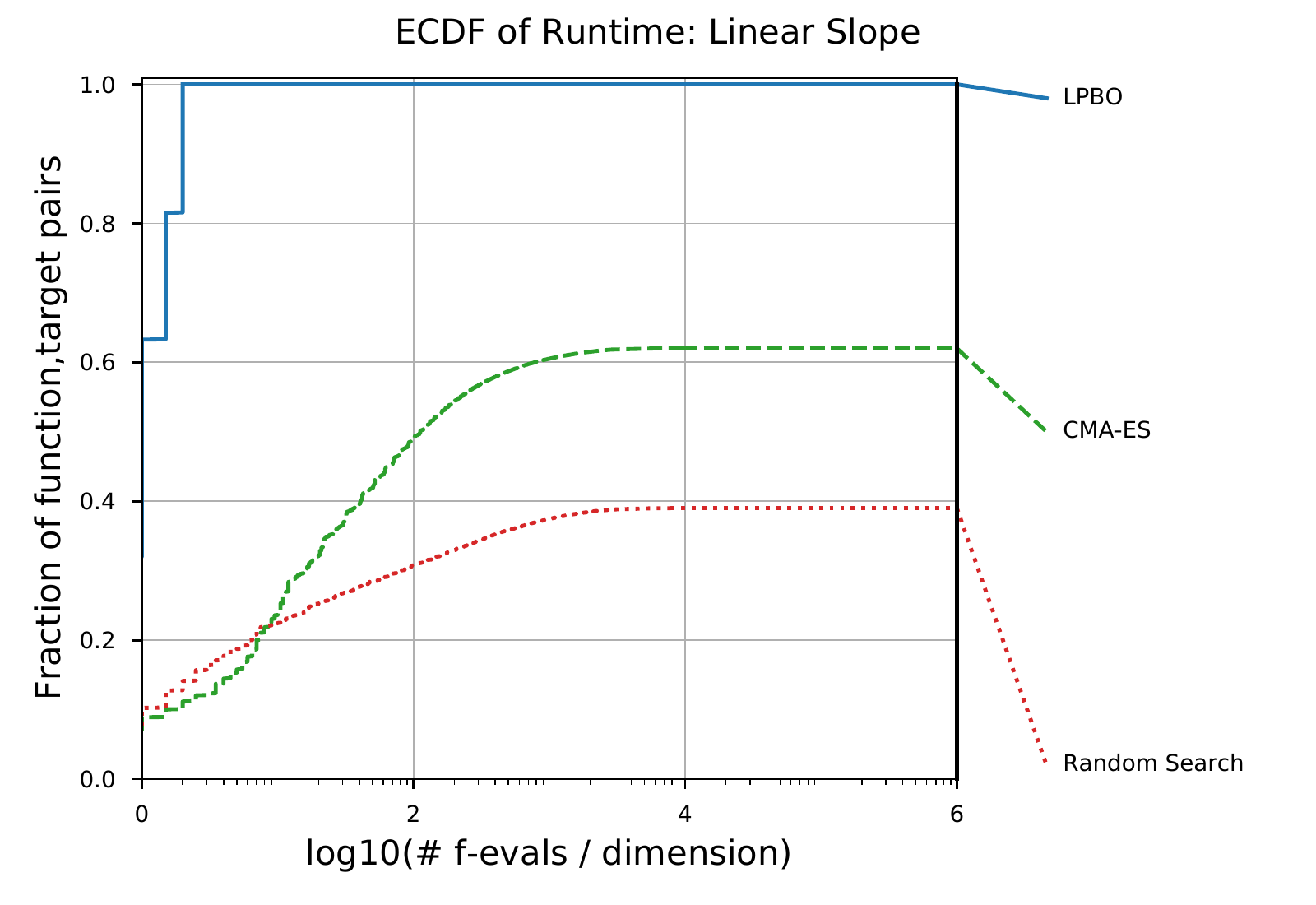}
 \end{subfigure}\hfil
 \begin{subfigure}{0.25\textwidth}
   \includegraphics[width=\linewidth]{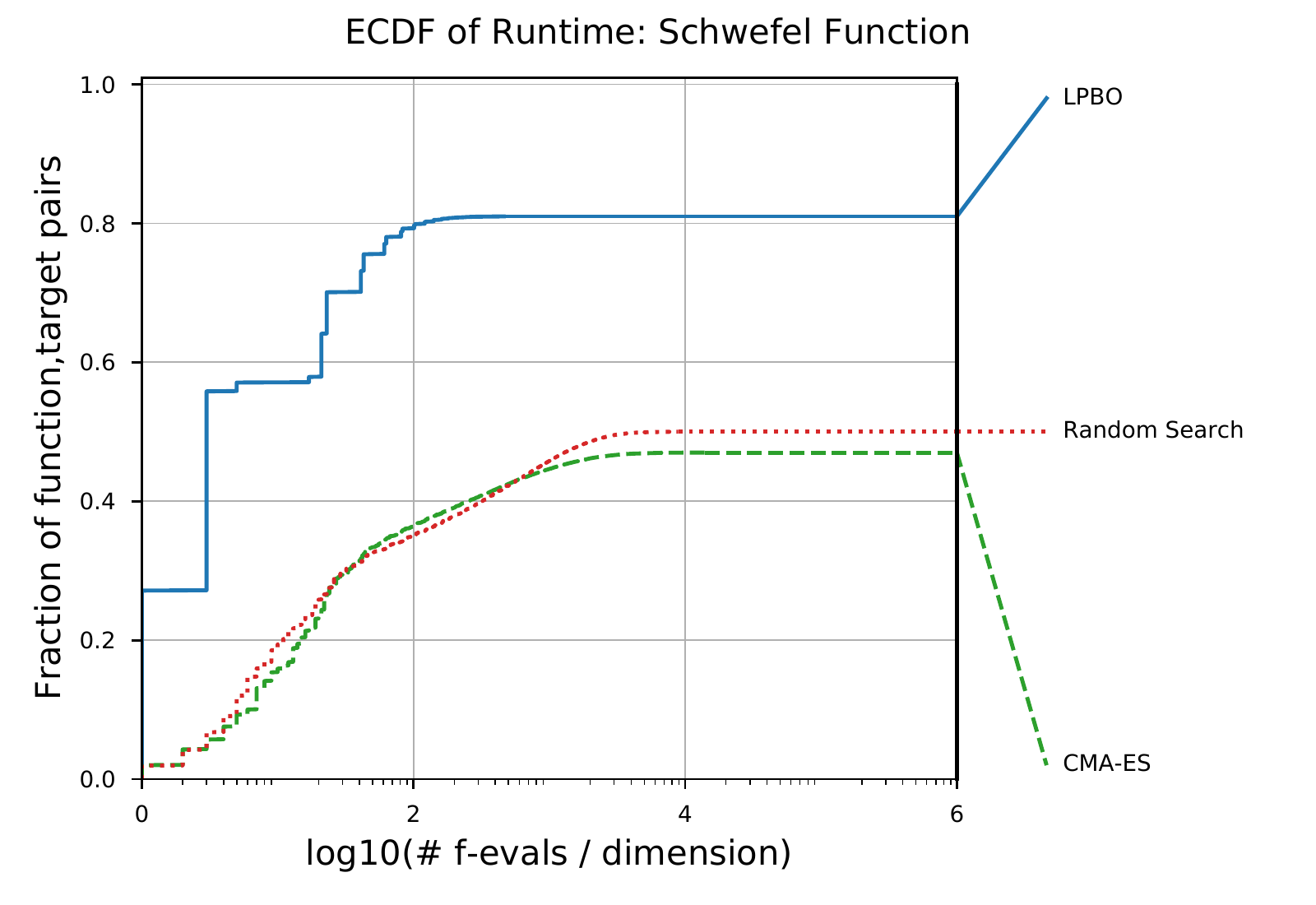}
 \end{subfigure}\hfil
 \begin{subfigure}{0.25\textwidth}
   \includegraphics[width=\linewidth]{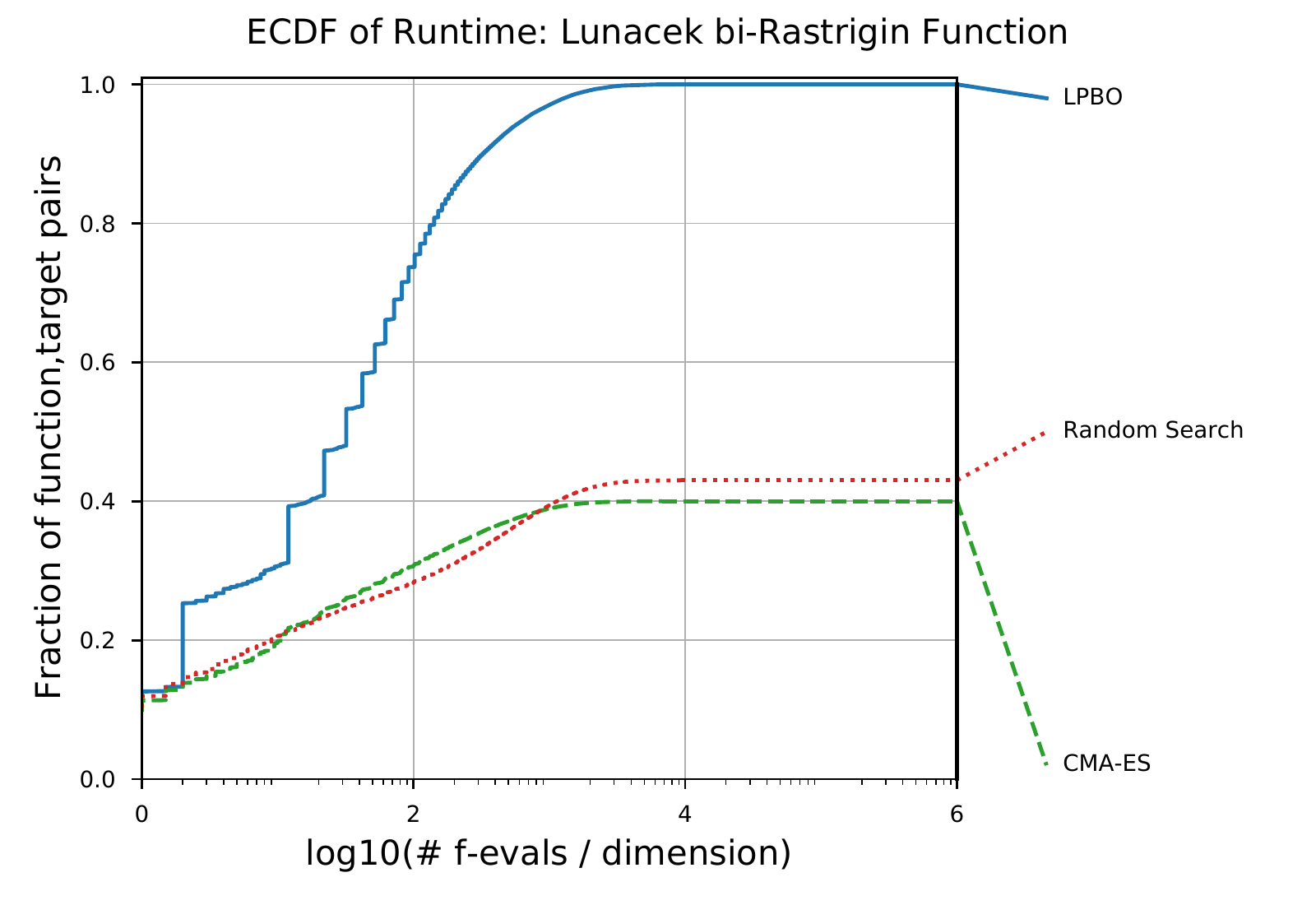}
 \end{subfigure}\hfil
 \begin{subfigure}{0.25\textwidth}
    \includegraphics[width=\linewidth]{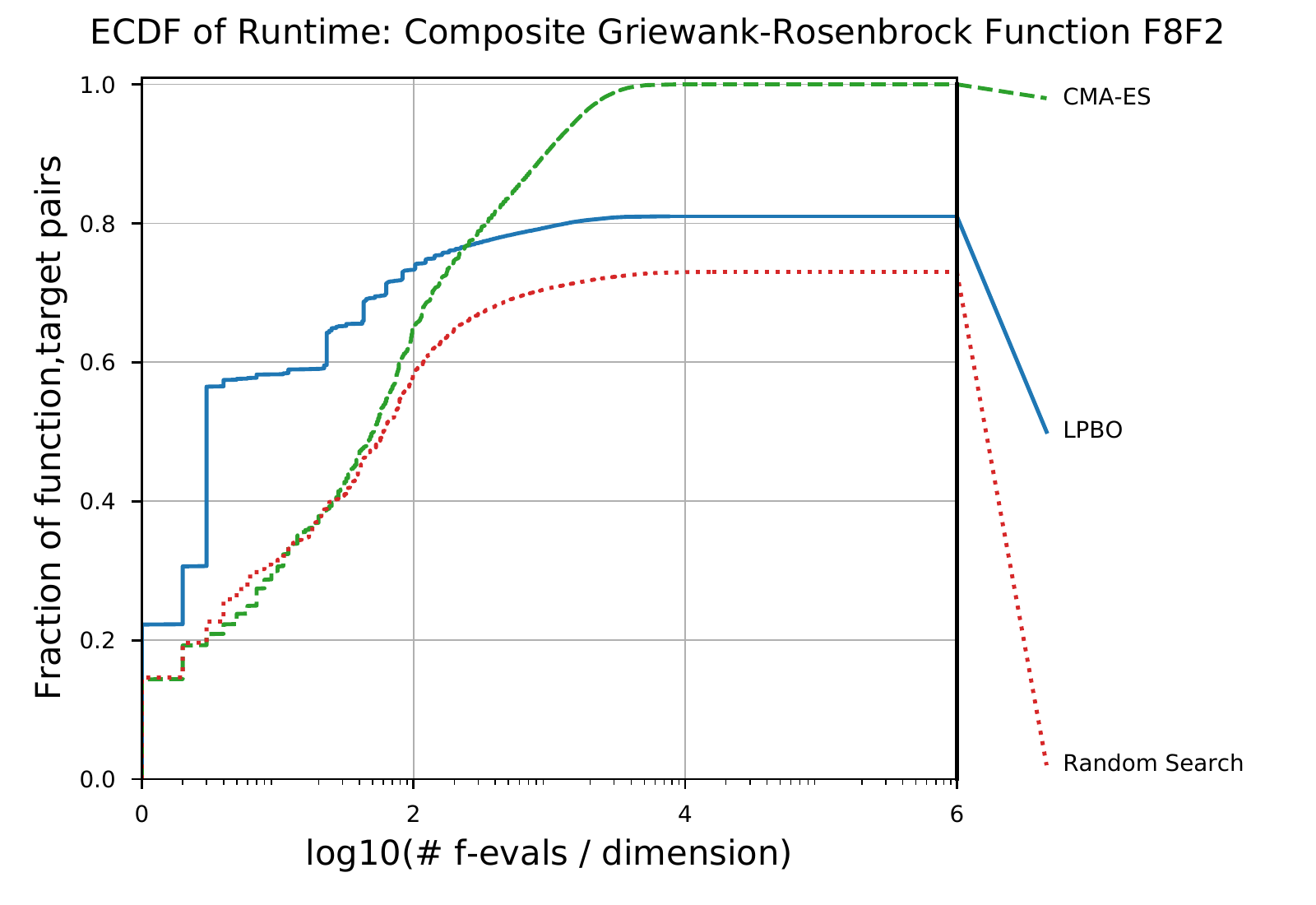}
 \end{subfigure}
 \begin{subfigure}{0.25\textwidth}
   \includegraphics[width=\linewidth]{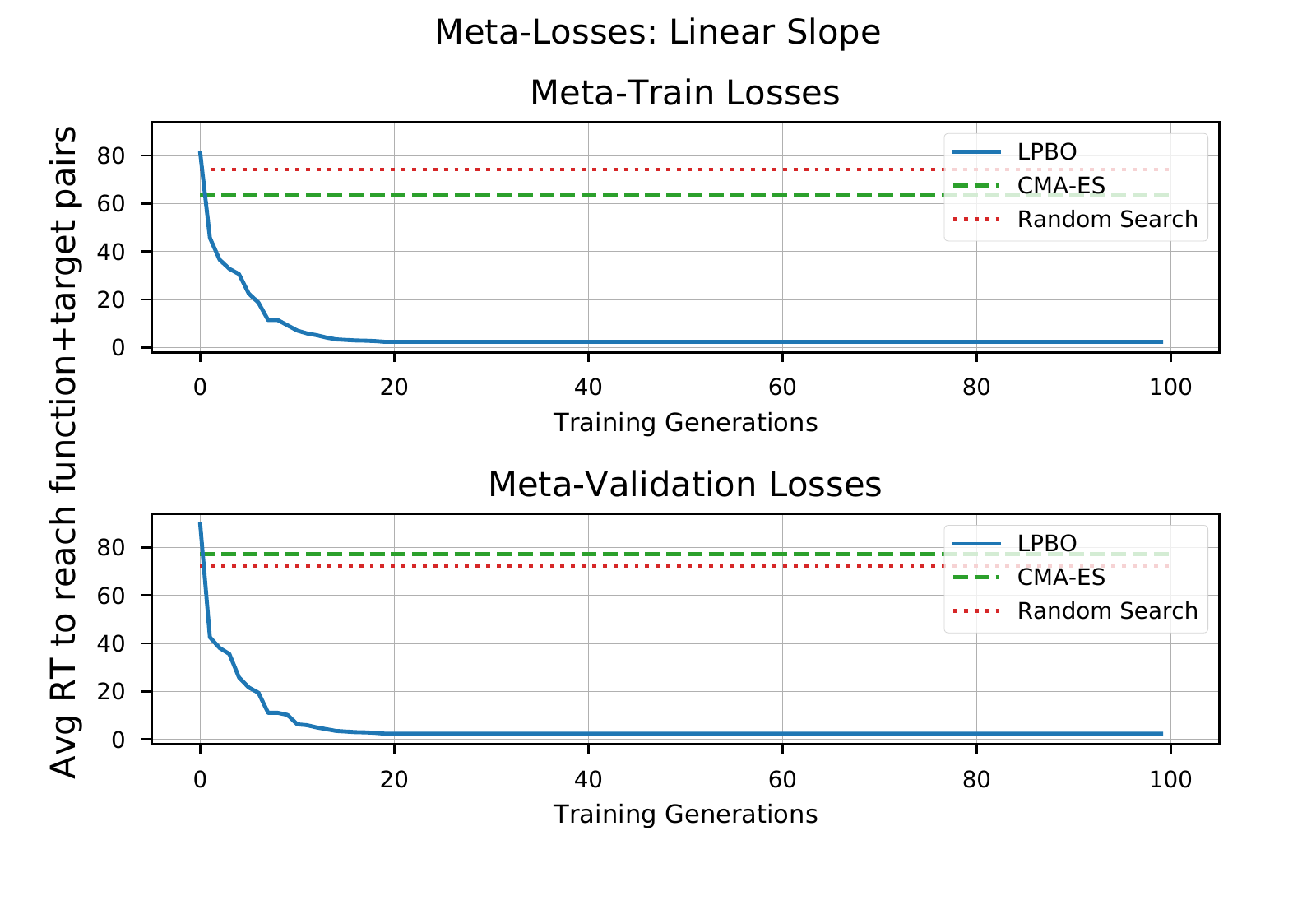}
 \end{subfigure}\hfil
 \begin{subfigure}{0.25\textwidth}
   \includegraphics[width=\linewidth]{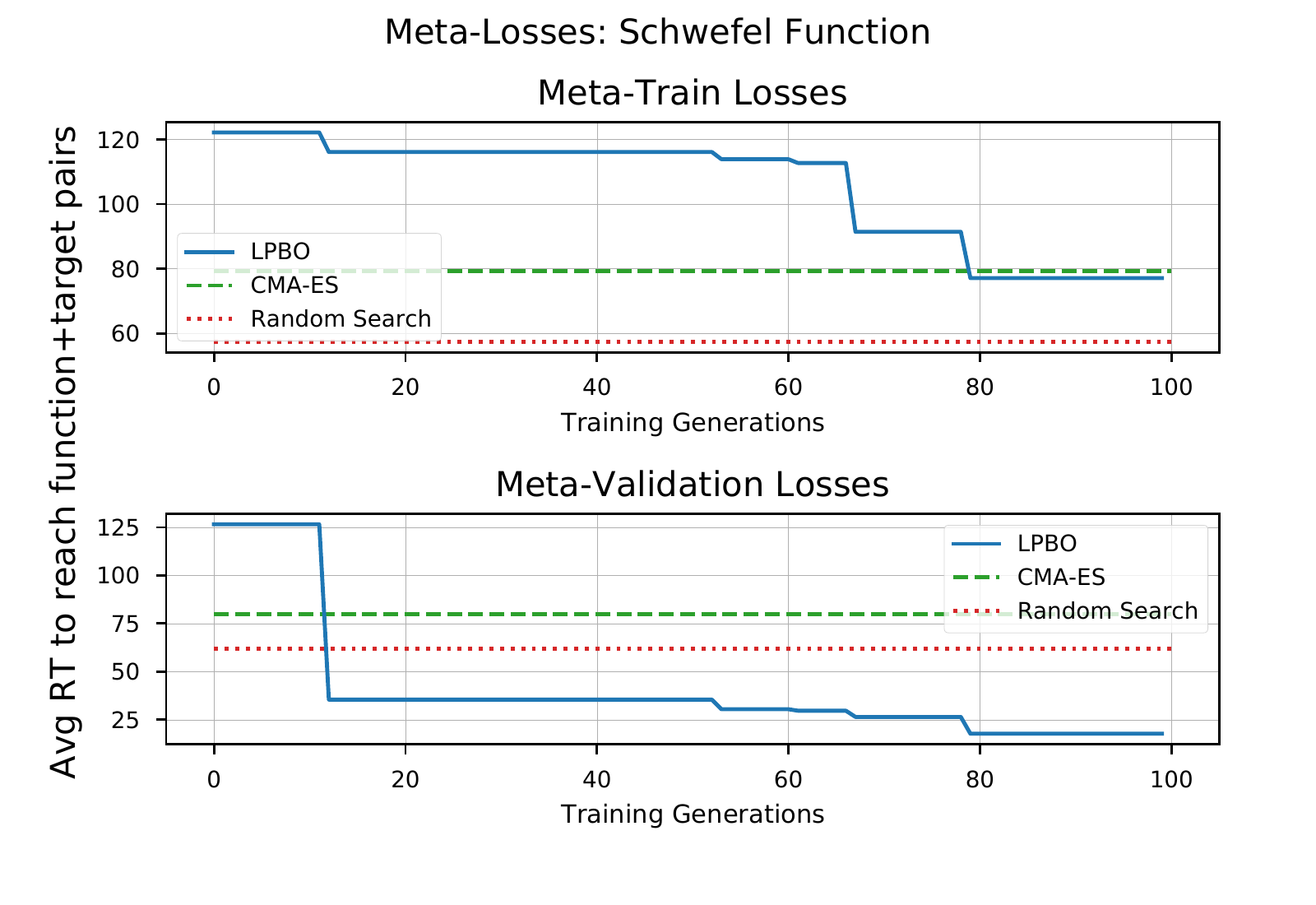}
 \end{subfigure}\hfil
 \begin{subfigure}{0.25\textwidth}
   \includegraphics[width=\linewidth]{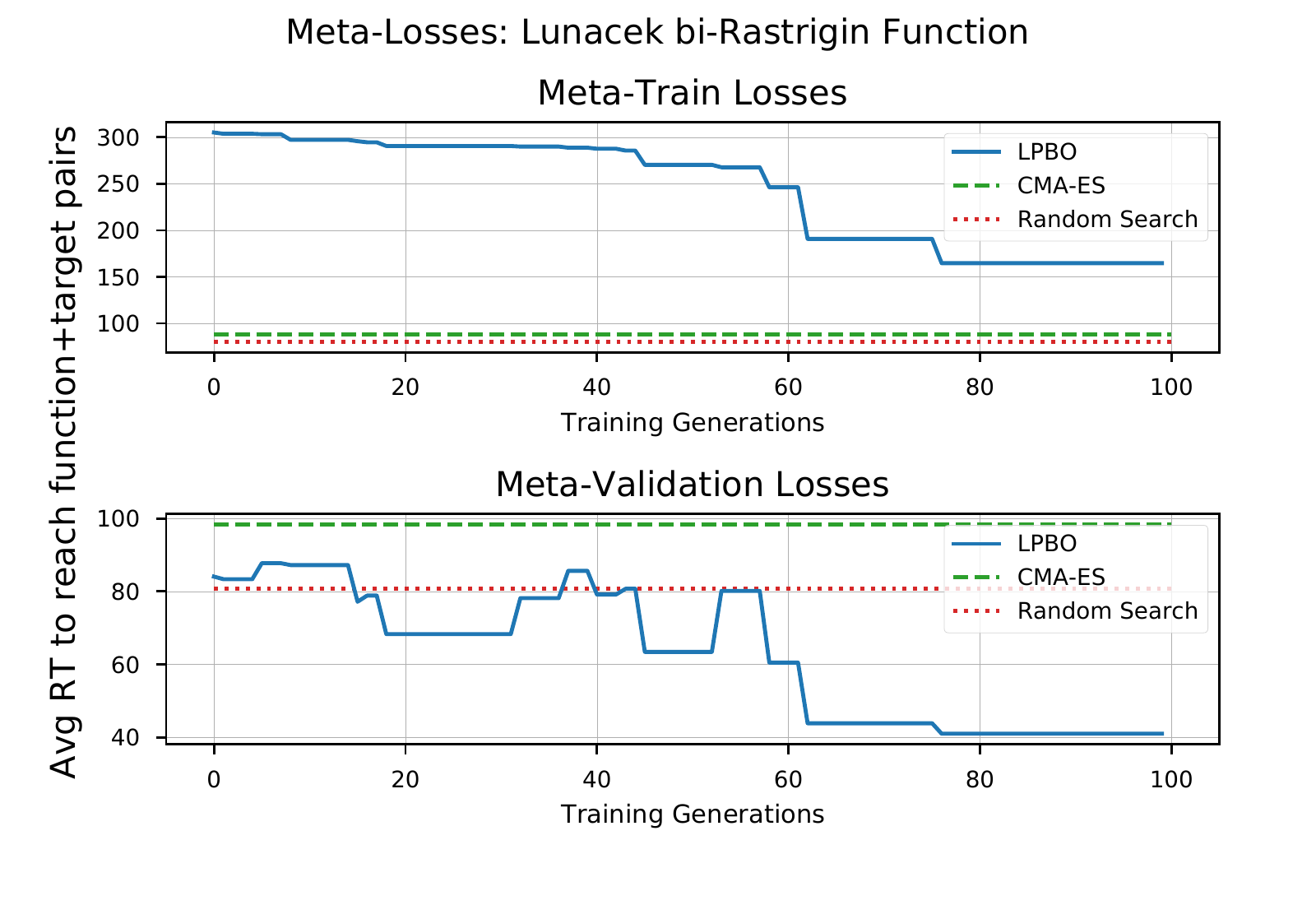}
 \end{subfigure}\hfil
 \begin{subfigure}{0.25\textwidth}
    \includegraphics[width=\linewidth]{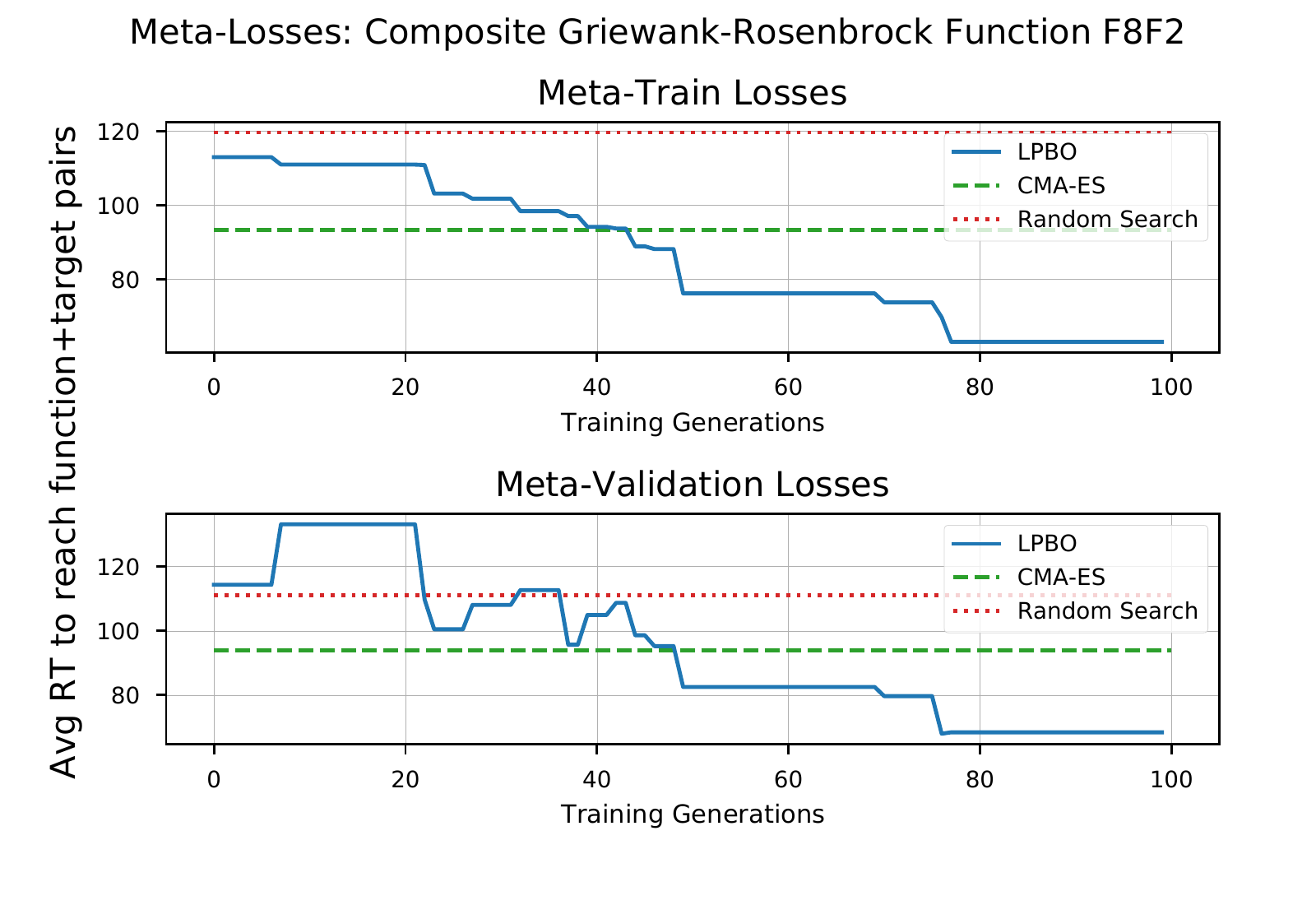}
\end{subfigure}
\caption{Results in meta-training multiple instances. ECDFs (top) and meta-losses (bottom) in \textit{Linear-Slope}, \textit{Schwefel},  \textit{Lunacek bi-Rastrigin} and \textit{Composite Griewank-Rosenbrock} functions in 2D (left to right).}
\label{fig:Experiments/BBOBSingle}
\end{figure*}
\subsubsection{Training and evaluation procedure rationales}

Two optimization benchmarks are chosen to be the training and evaluation dataset: the COCO platform \cite{hansen2020coco}, and machine learning algorithm's hyperparameter optimization (HPO) \cite{hpo_citation}. Both provide an interface to generate a wide range of related functions, and we can aggregate them in a class of problems. From these, we create 1000 different instances of each function using 100 as training, 100 as validation, and 800 as testing. This results in training the policy $100 N$ times if $N$ different functions have been selected. We wrapped the COCO and HPO interface to our POMDP implementation, consistent with the OpenAI Gym \cite{brockman2016openai}. As baselines, we have Random Search (RS), a batch-version of random search where every iteration of this algorithm is sampling independent search points uniformly from the search space, and Covariance matrix adaptation evolution strategy (CMA-ES) \cite{hansen_cma_2016}, often considered as the state-of-the-art method for continuous domain optimization under challenging settings (e.g., ill-conditioned, non-convex, non-continuous, multimodal).

\begin{figure}[!htb]
    \centering
\begin{subfigure}{0.15\textwidth}
  \includegraphics[width=\linewidth]{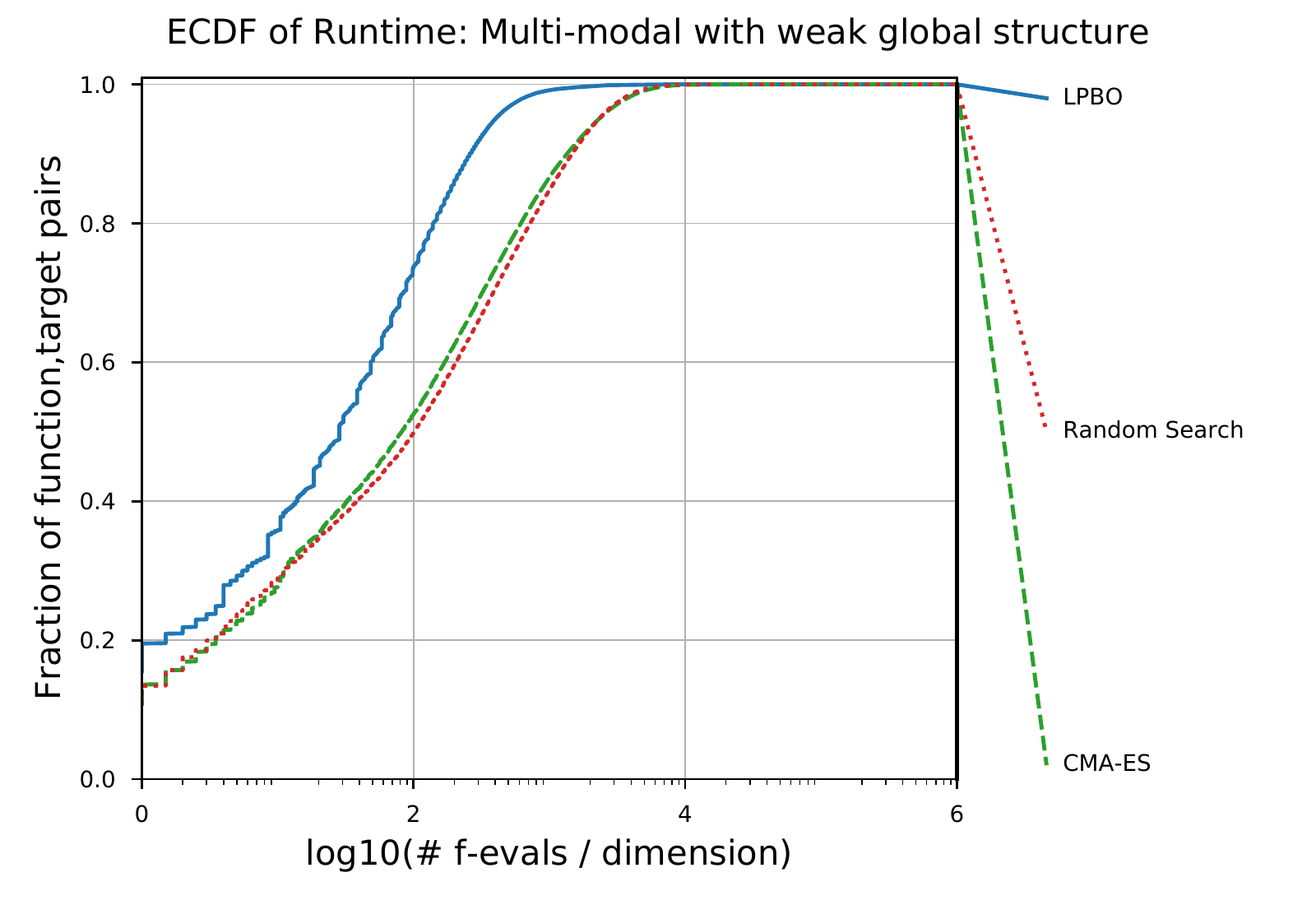}
\end{subfigure}\hfil
\begin{subfigure}{0.15\textwidth}
  \includegraphics[width=\linewidth]{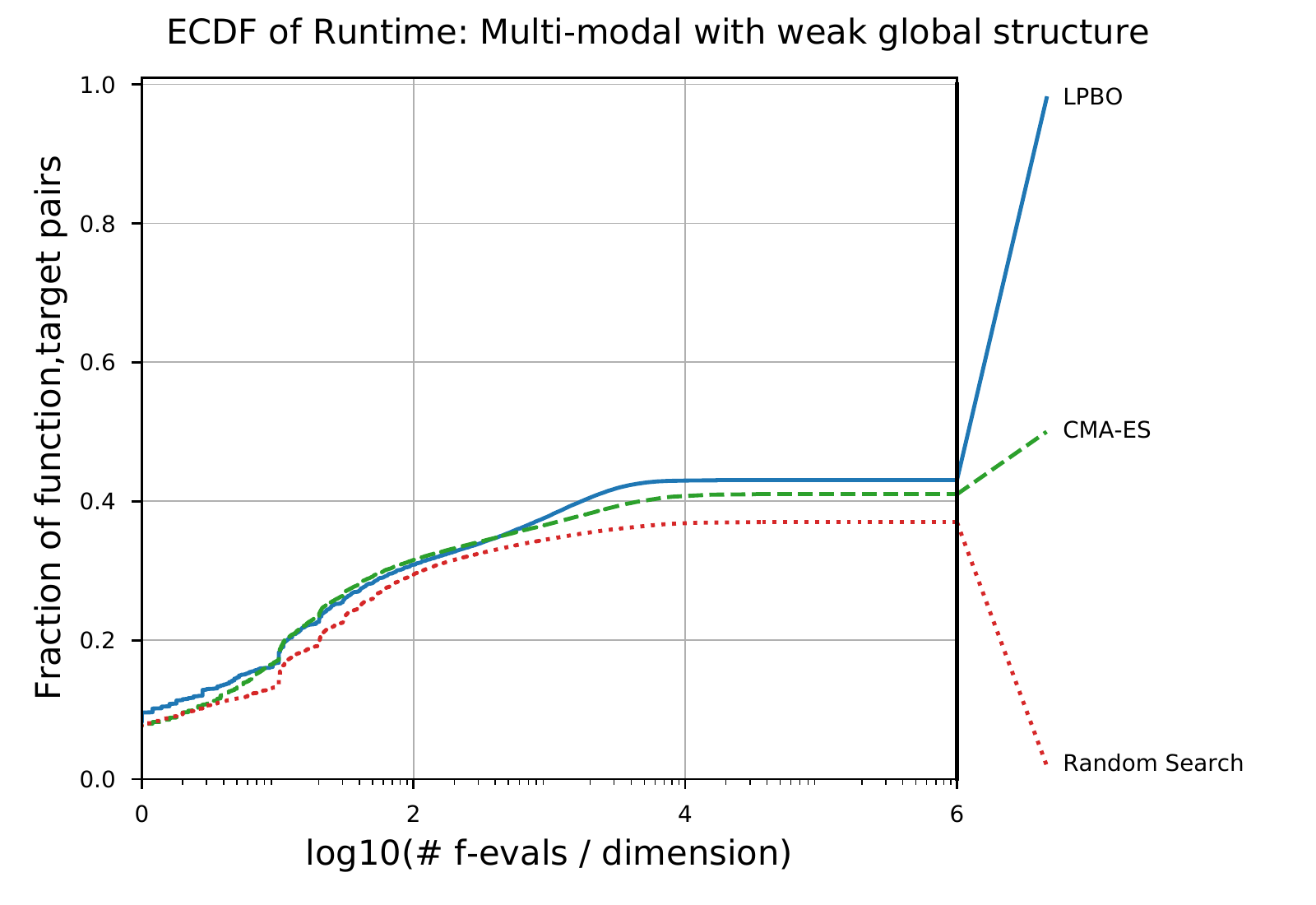}
\end{subfigure}
\begin{subfigure}{0.15\textwidth}
  \includegraphics[width=\linewidth]{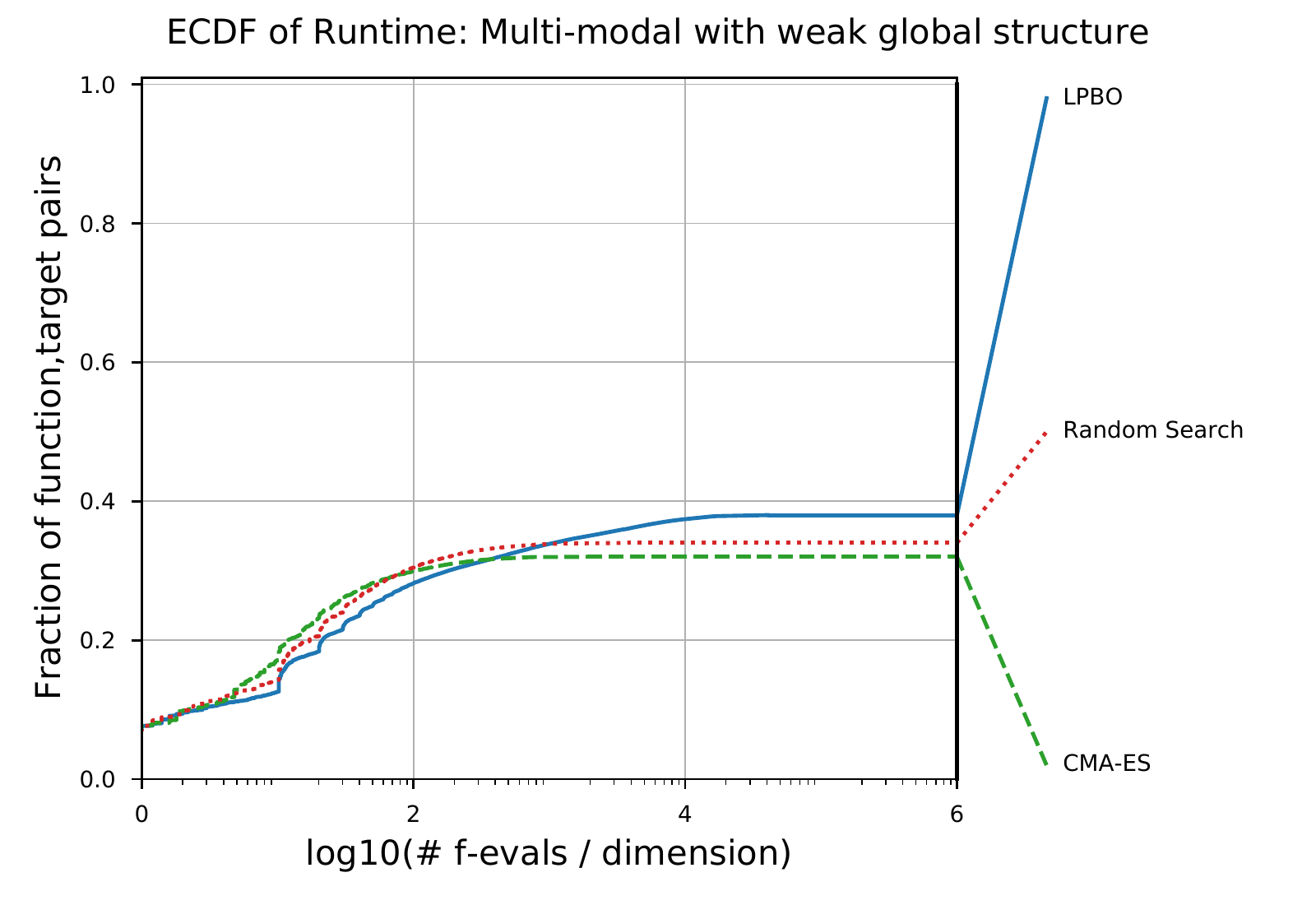}
\end{subfigure}

\begin{subfigure}{0.15\textwidth}
  \includegraphics[width=\linewidth]{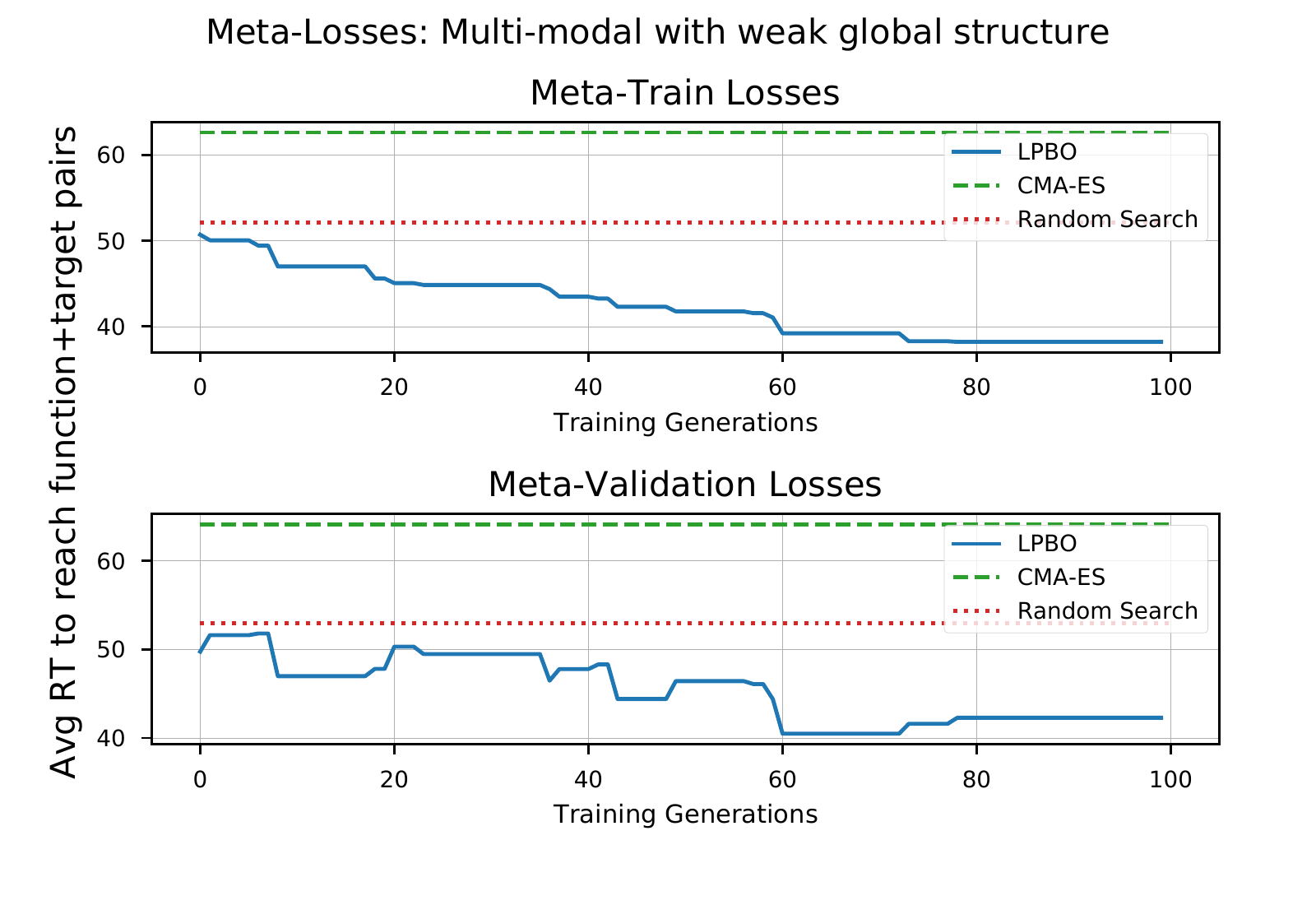}
\end{subfigure}\hfil
\begin{subfigure}{0.15\textwidth}
  \includegraphics[width=\linewidth]{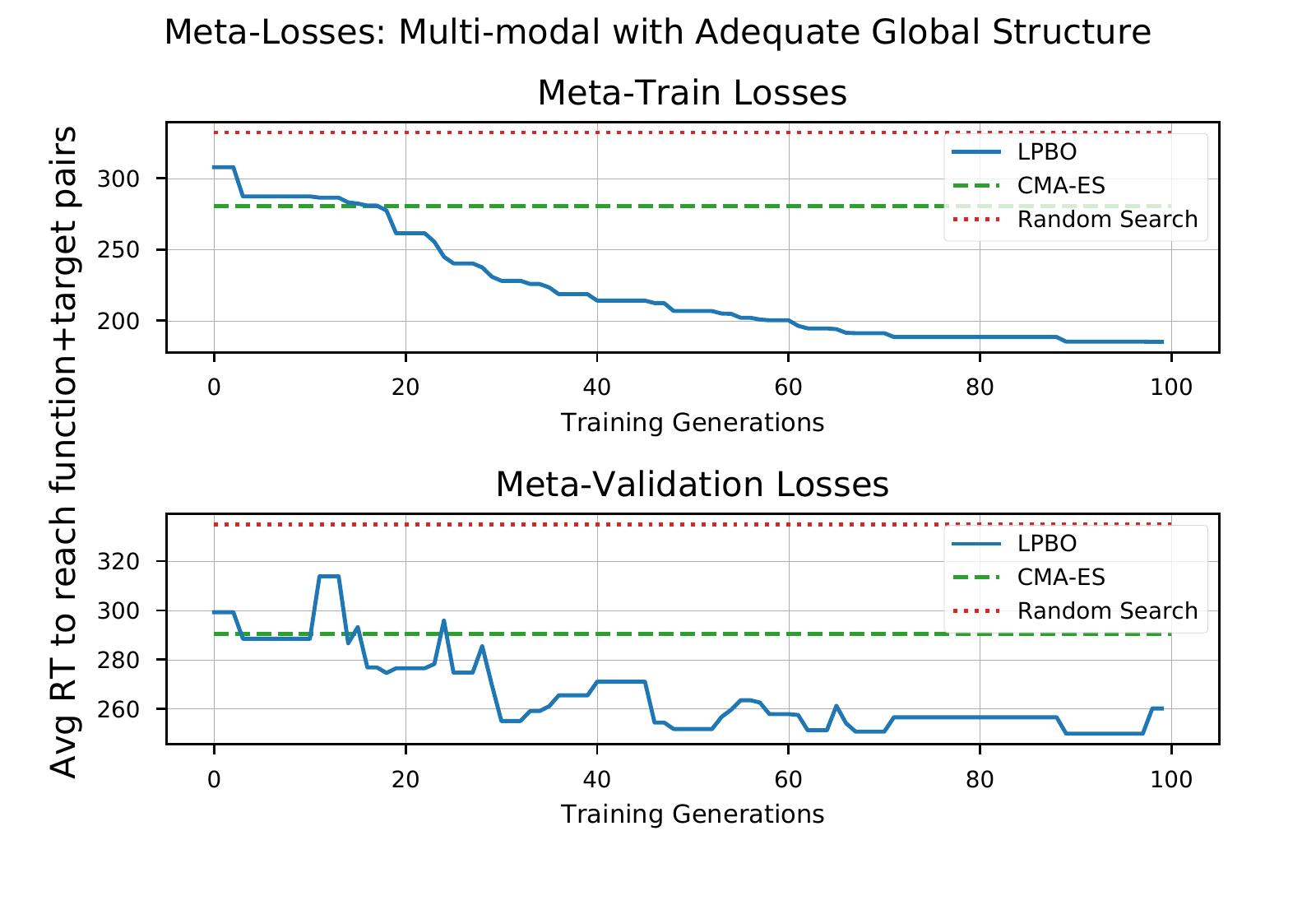}
\end{subfigure}
\begin{subfigure}{0.15\textwidth}
  \includegraphics[width=\linewidth]{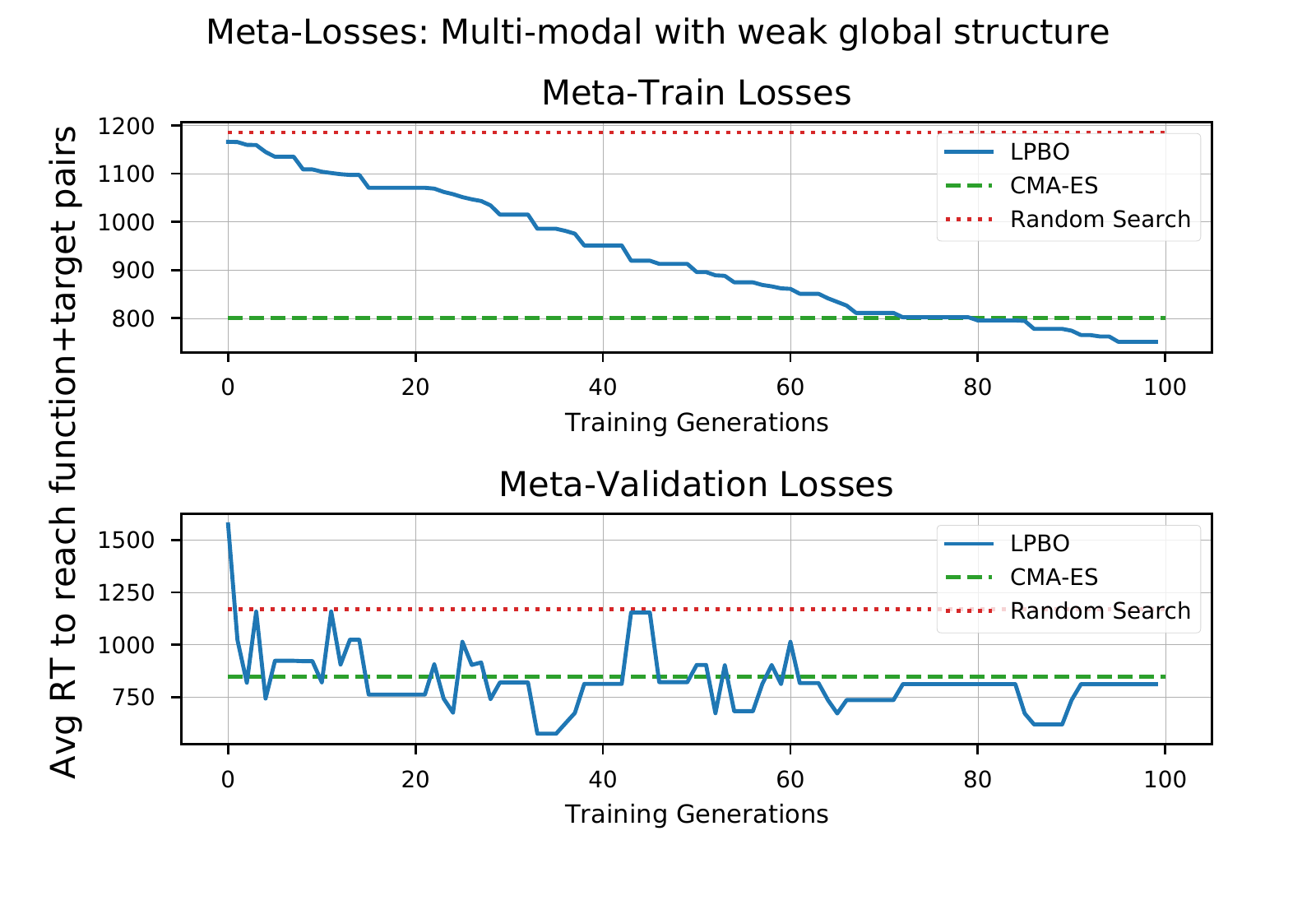}
\end{subfigure}

\begin{subfigure}{0.15\textwidth}
  \includegraphics[width=\linewidth]{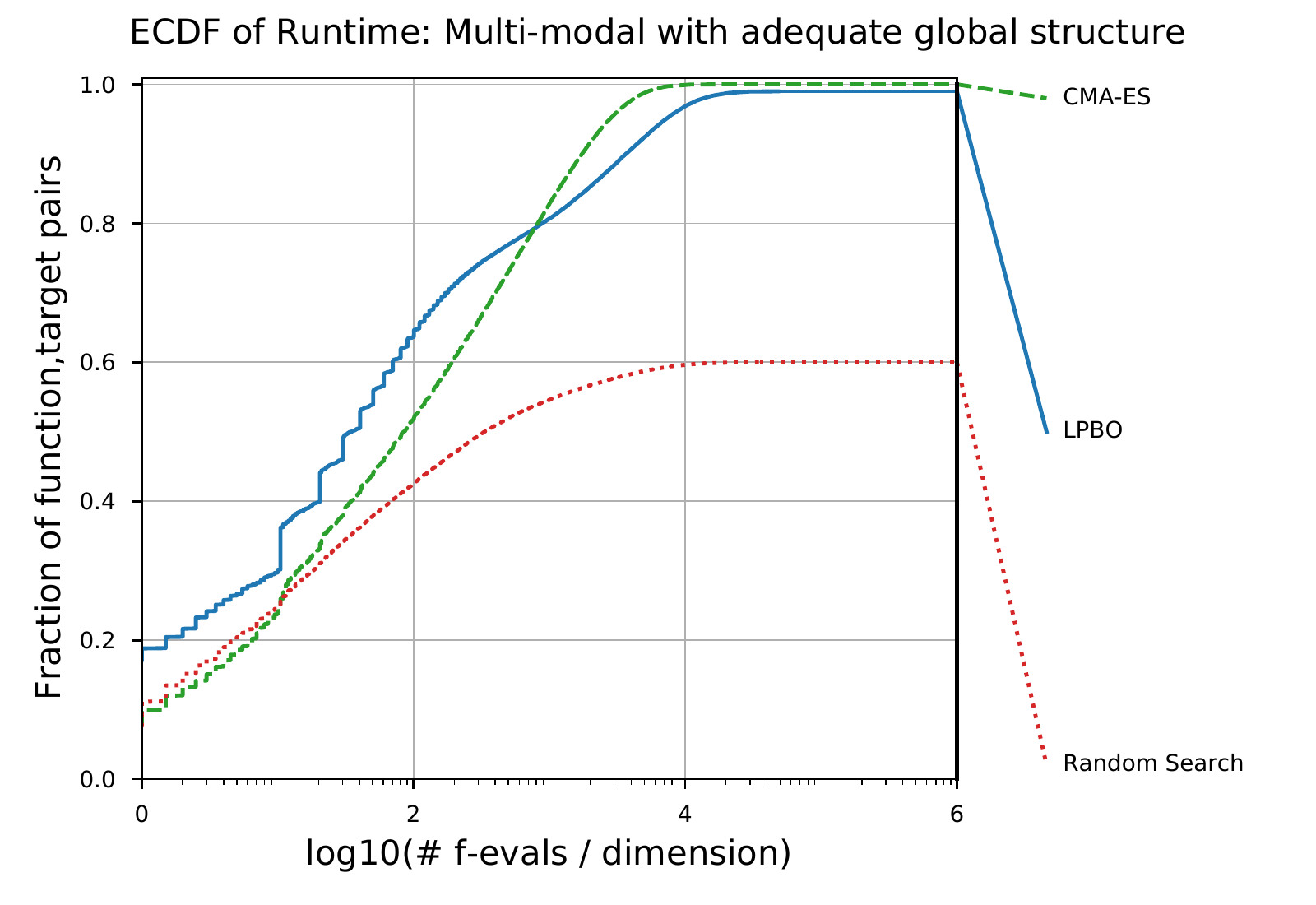}
\end{subfigure}\hfil
\begin{subfigure}{0.15\textwidth}
  \includegraphics[width=\linewidth]{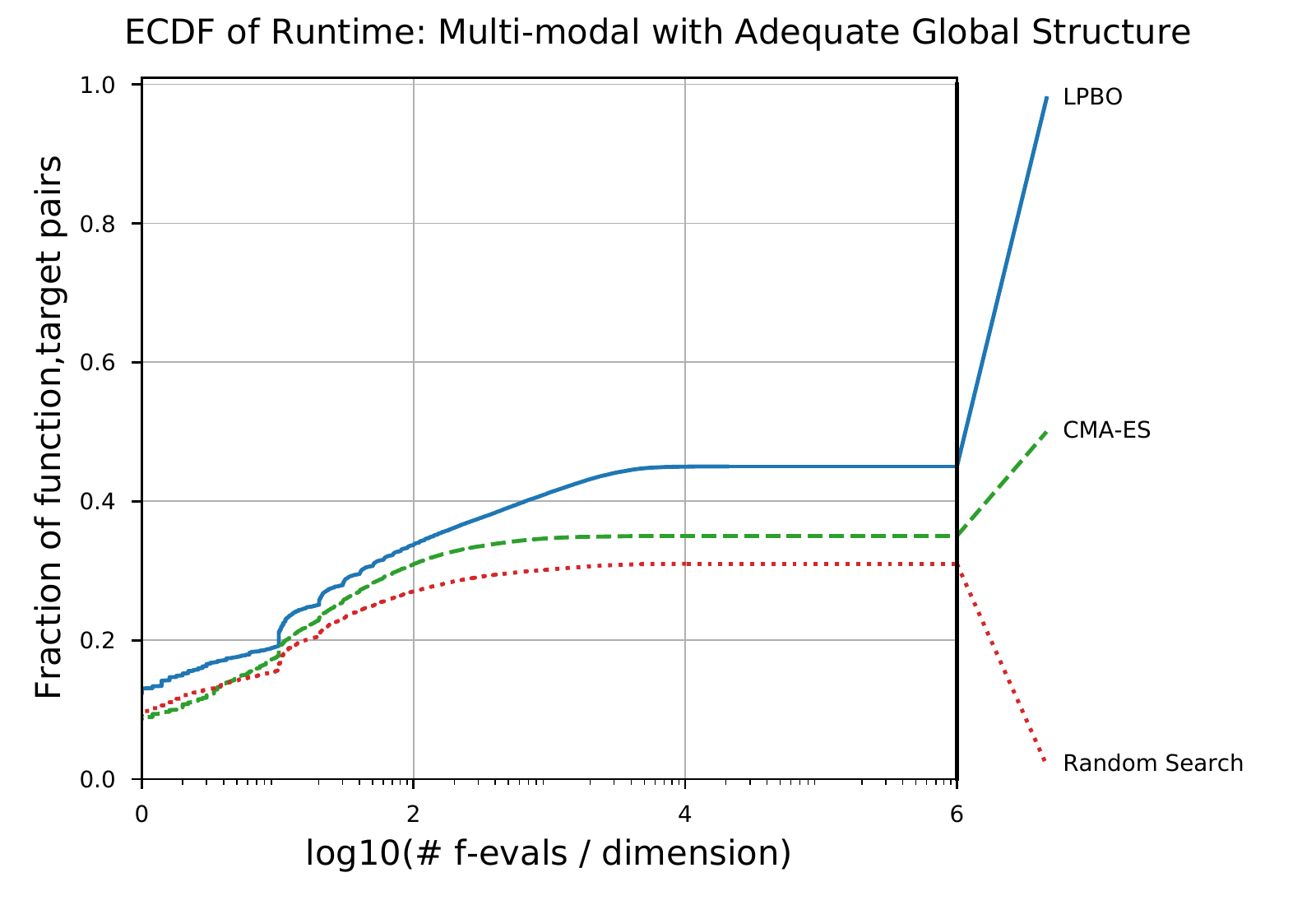}
\end{subfigure}
\begin{subfigure}{0.15\textwidth}
  \includegraphics[width=\linewidth]{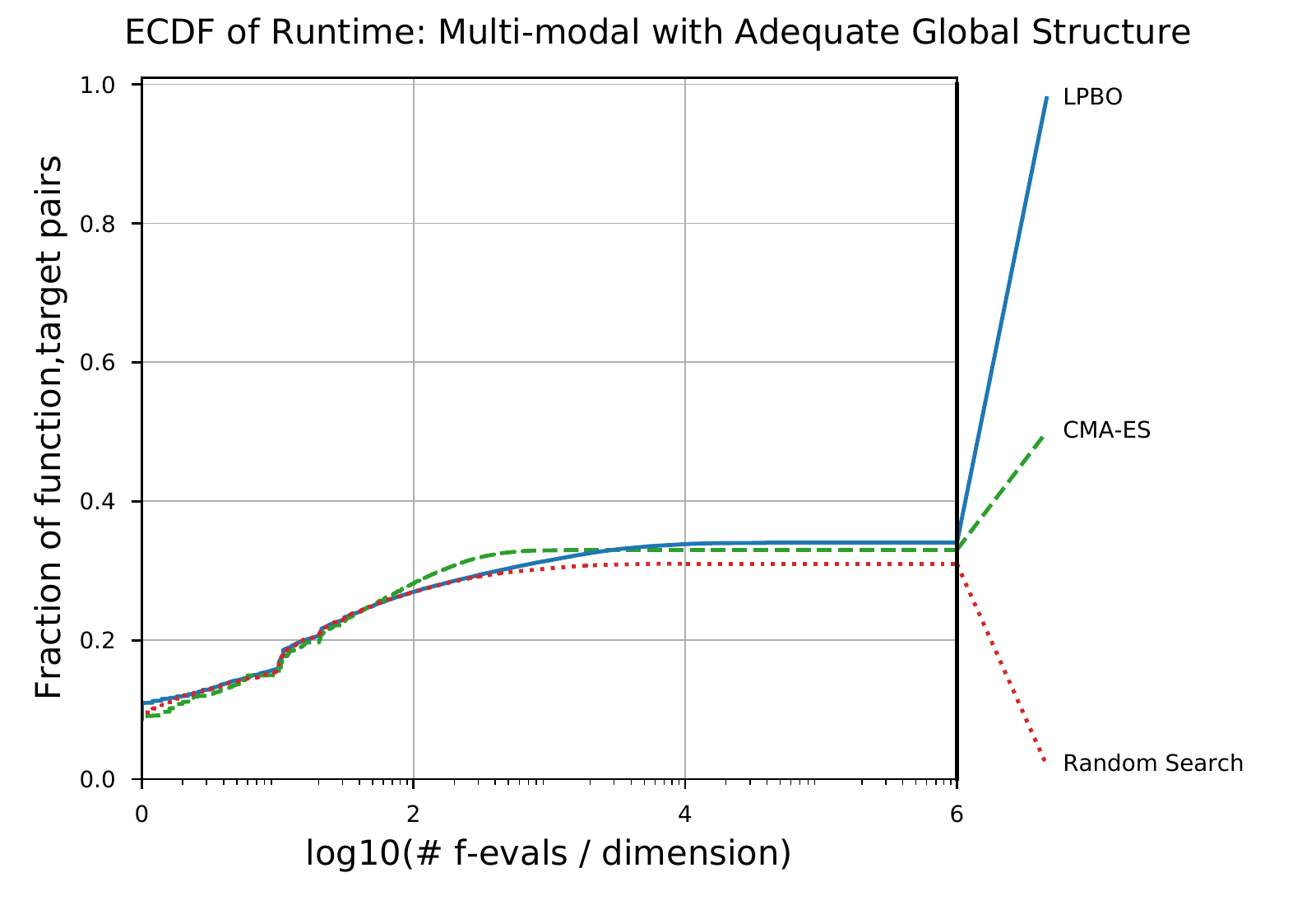}
\end{subfigure}

\begin{subfigure}{0.15\textwidth}
  \includegraphics[width=\linewidth]{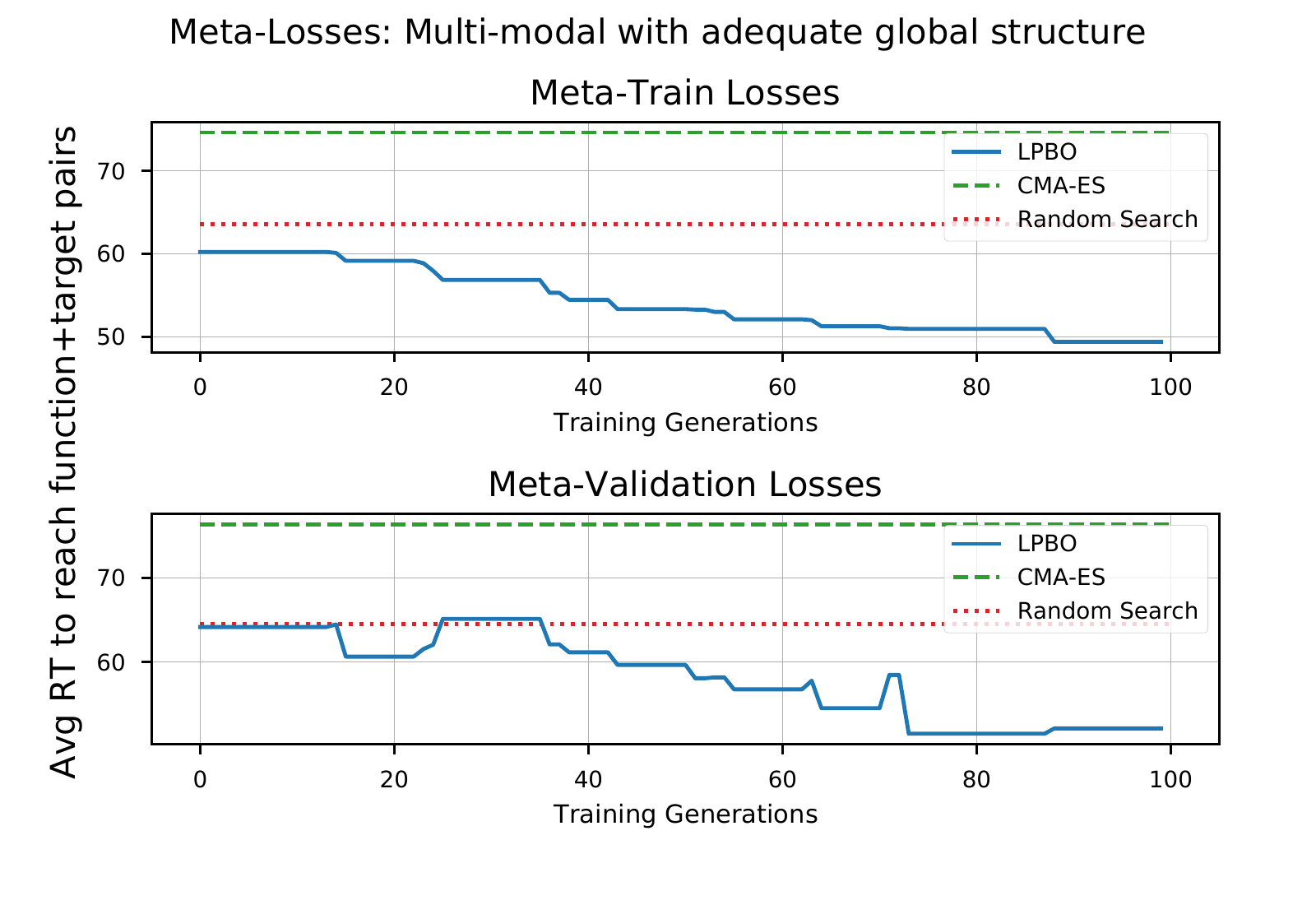}
\end{subfigure}\hfil
\begin{subfigure}{0.15\textwidth}
  \includegraphics[width=\linewidth]{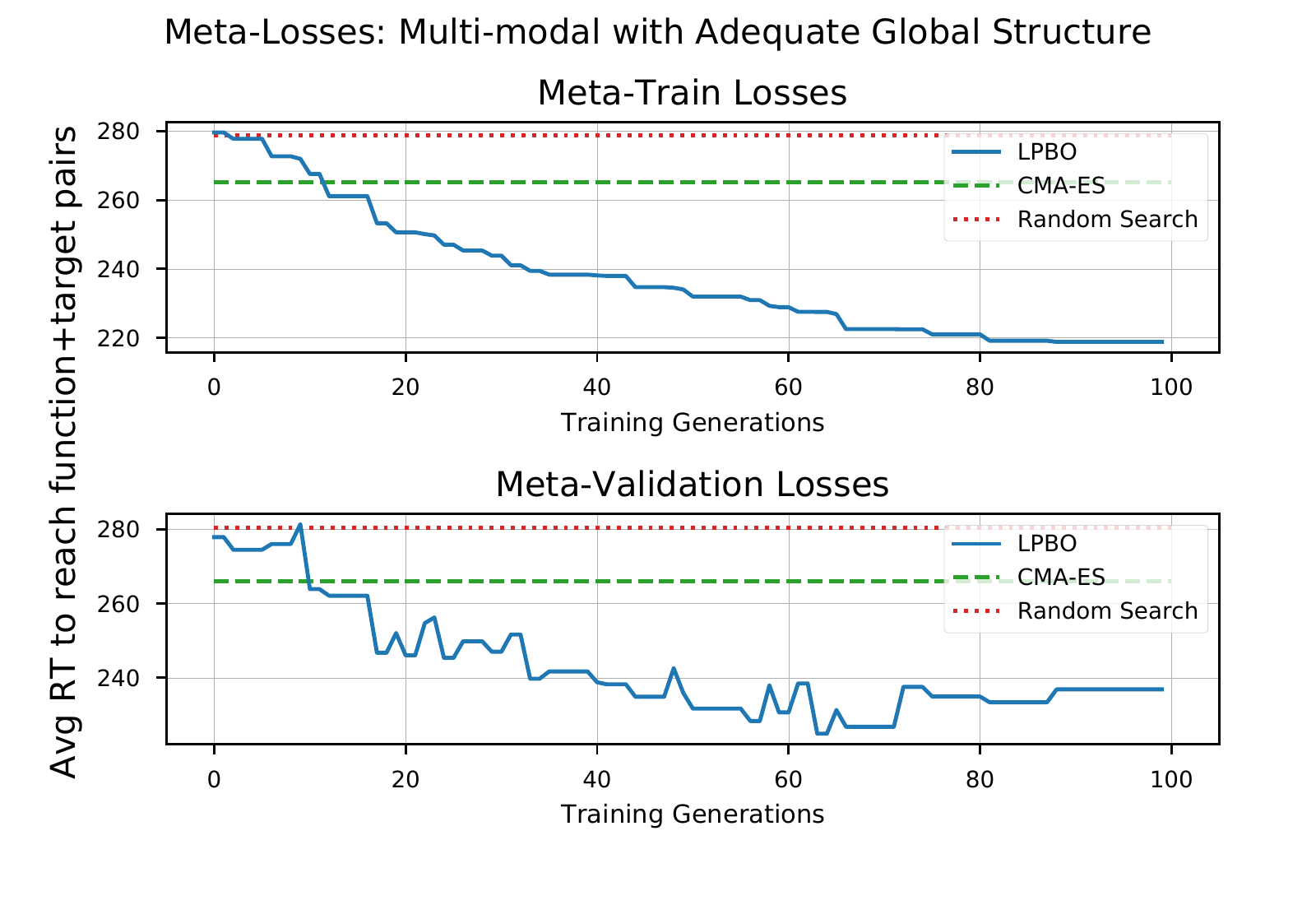}
\end{subfigure}
\begin{subfigure}{0.15\textwidth}
  \includegraphics[width=\linewidth]{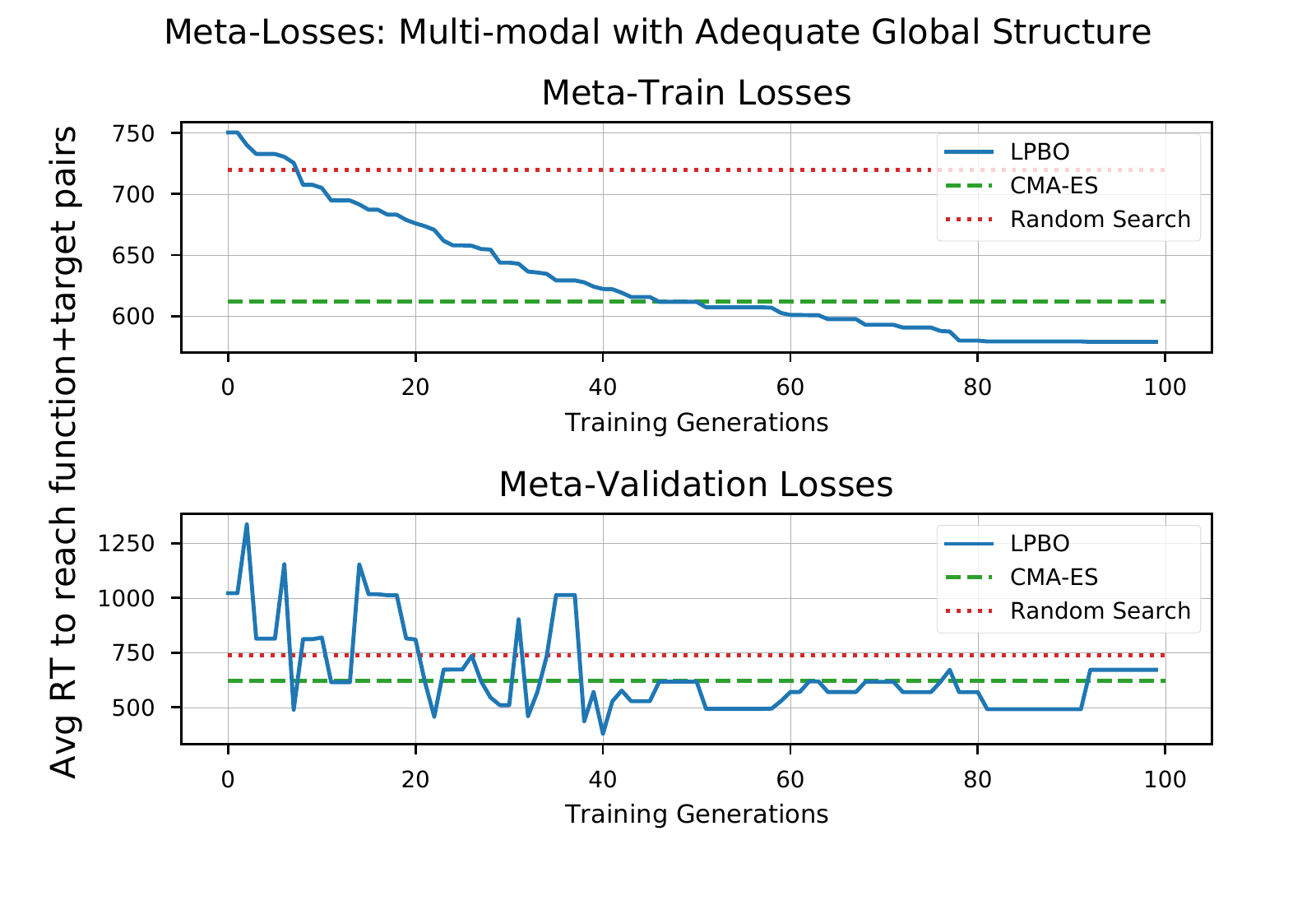}
\end{subfigure}
\caption{Results in meta-training multiple functions. ECDFs and meta-losses for group 1 (top) and group 2 (bottom) in 2D, 5D and 10D (left to right).}
\label{fig:Experiments/BBOBGroup}
\end{figure}

We proceed as follows to evaluate and compare the baselines with the learned population-based optimizer (LPBO). To access the meta-learning, we show a meta-loss plot for training and validation tasks (see Fig.~\ref{fig:Experiments/BBOBGroup}). This plot measures performance and generalization over a distribution of tasks, with the meta-objective described in Sec.~\ref{sec:4.2} on the y-axis. We present the value for the best individual of the meta-optimizer population over the training tasks. From an optimization perspective, we are looking at the performance of an inner-optimization loop. We use the same benchmarking procedure on the COCO platform for that purpose, with the difference that tolerance of $10^{-3}$ to the global optimum value and $100 * D$ evaluations are considered. We aggregate the results using the test dataset, which was not accessible to the learned optimizer during its meta-training.

\begin{figure*}[!htb]
    \centering
\begin{subfigure}{0.3\textwidth}
  \includegraphics[width=\linewidth]{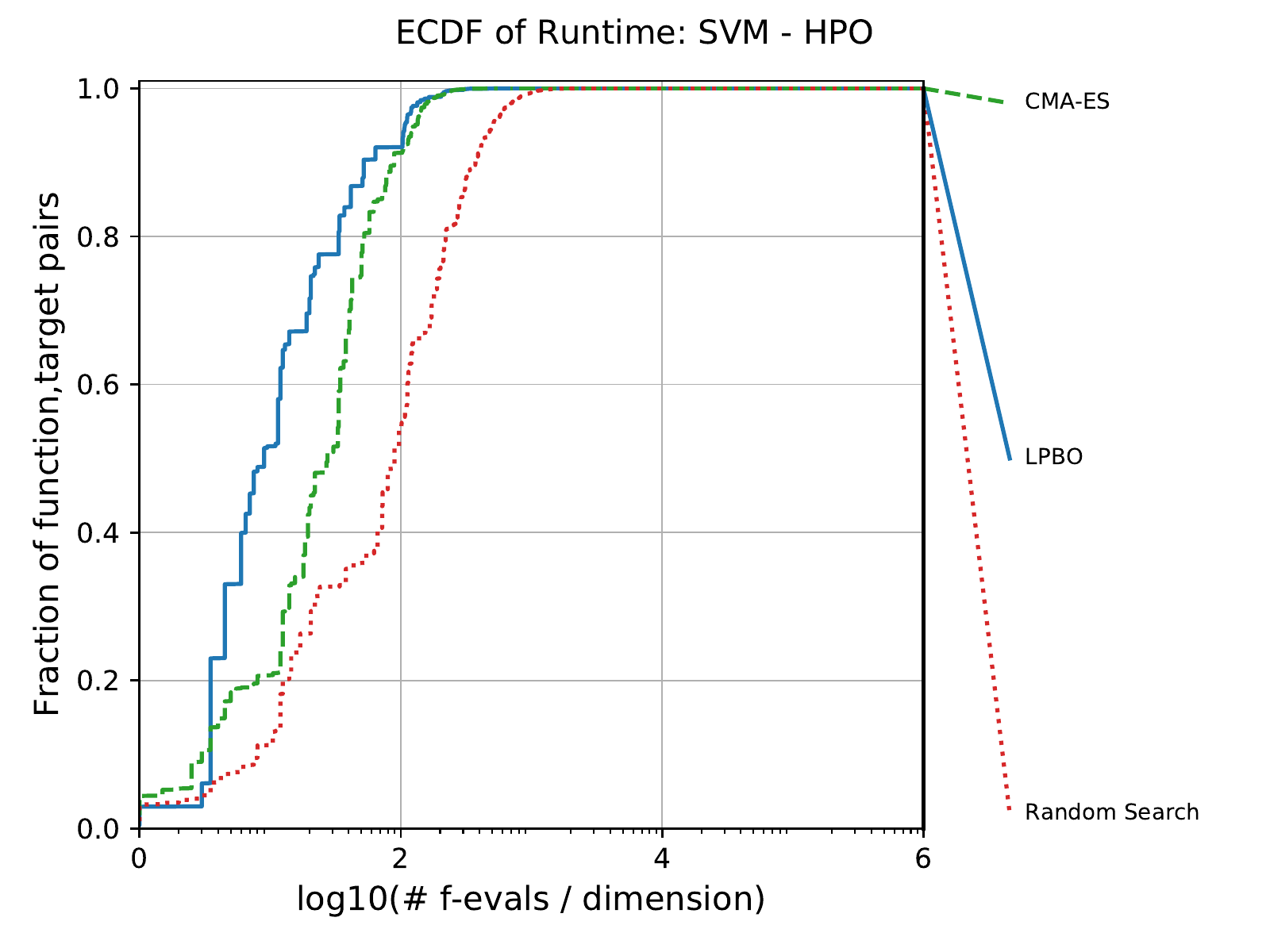}
\end{subfigure}\hfil
\begin{subfigure}{0.3\textwidth}
  \includegraphics[width=\linewidth]{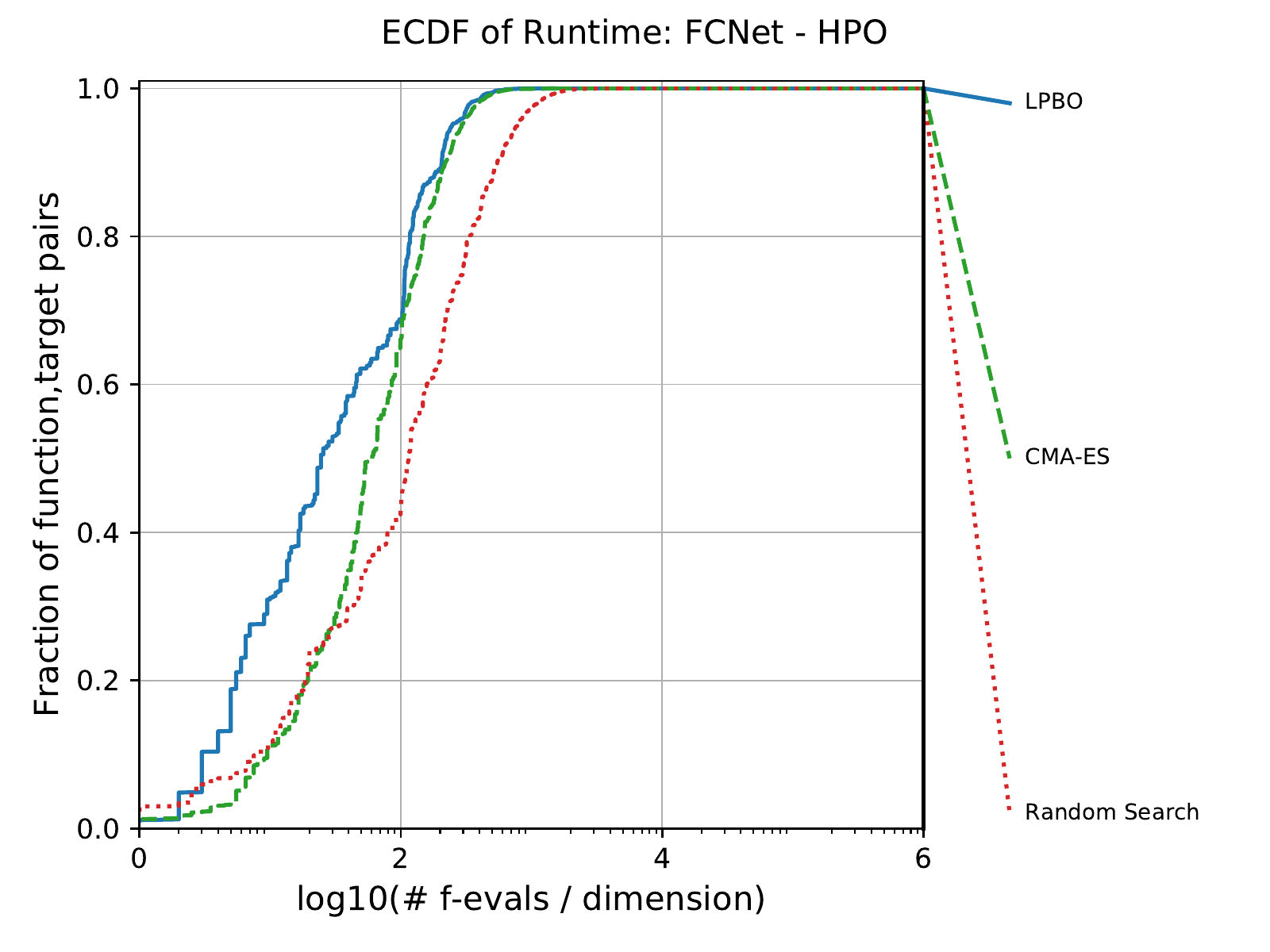}
\end{subfigure}\hfil
\begin{subfigure}{0.3\textwidth}
  \includegraphics[width=\linewidth]{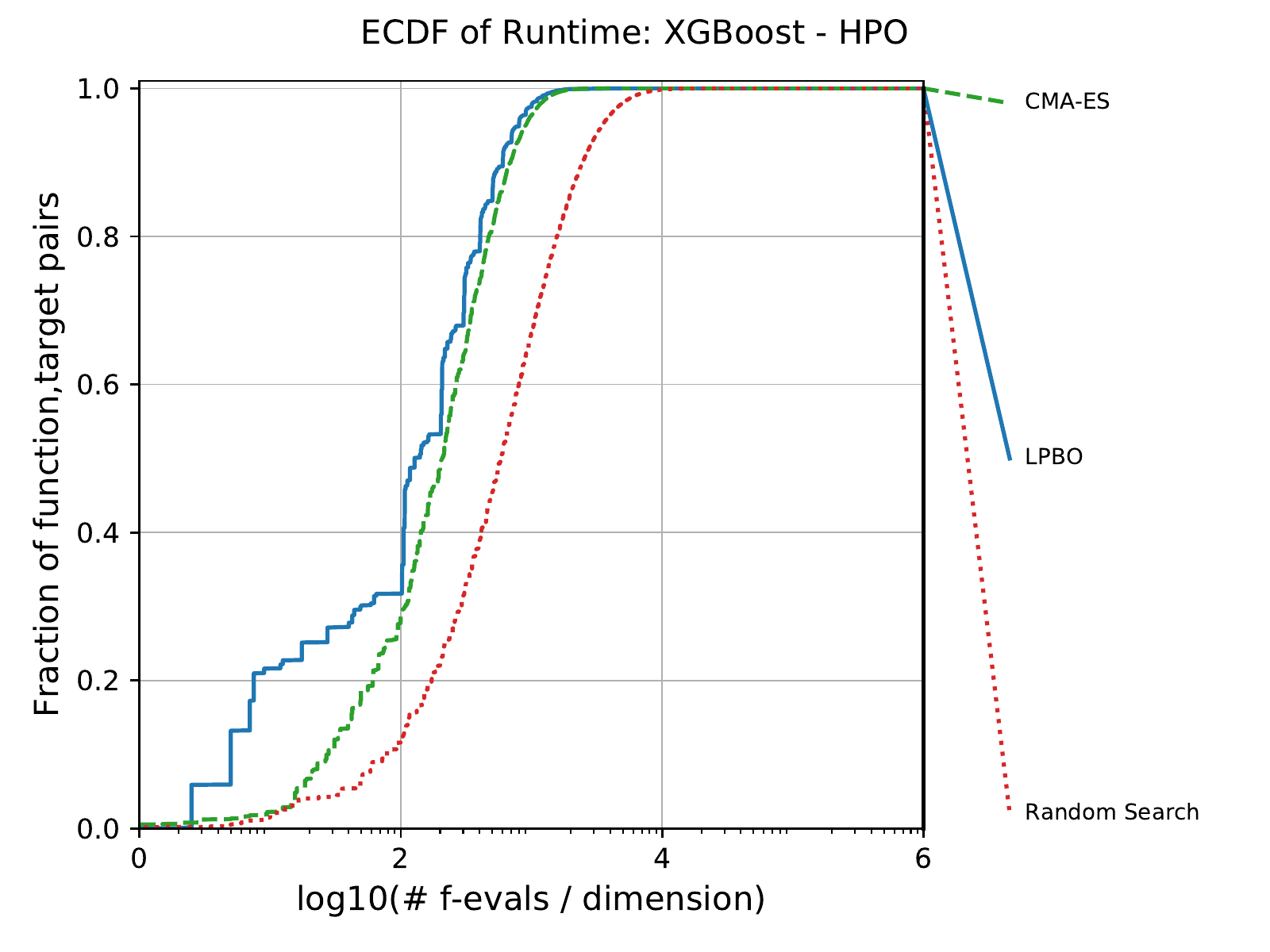}
\end{subfigure}

\begin{subfigure}{0.3\textwidth}
  \includegraphics[width=\linewidth]{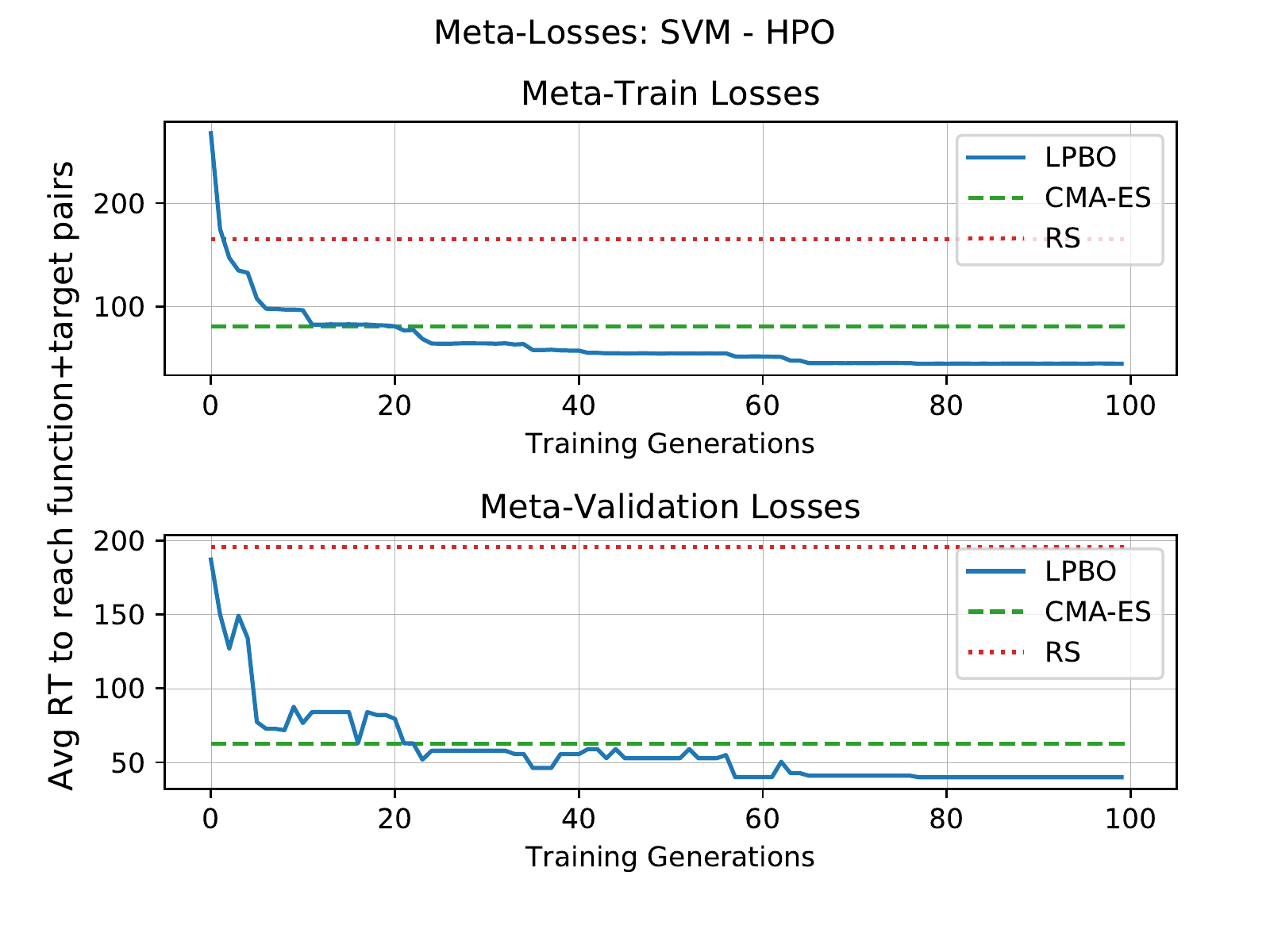}
\end{subfigure}\hfil 
\begin{subfigure}{0.3\textwidth}
  \includegraphics[width=\linewidth]{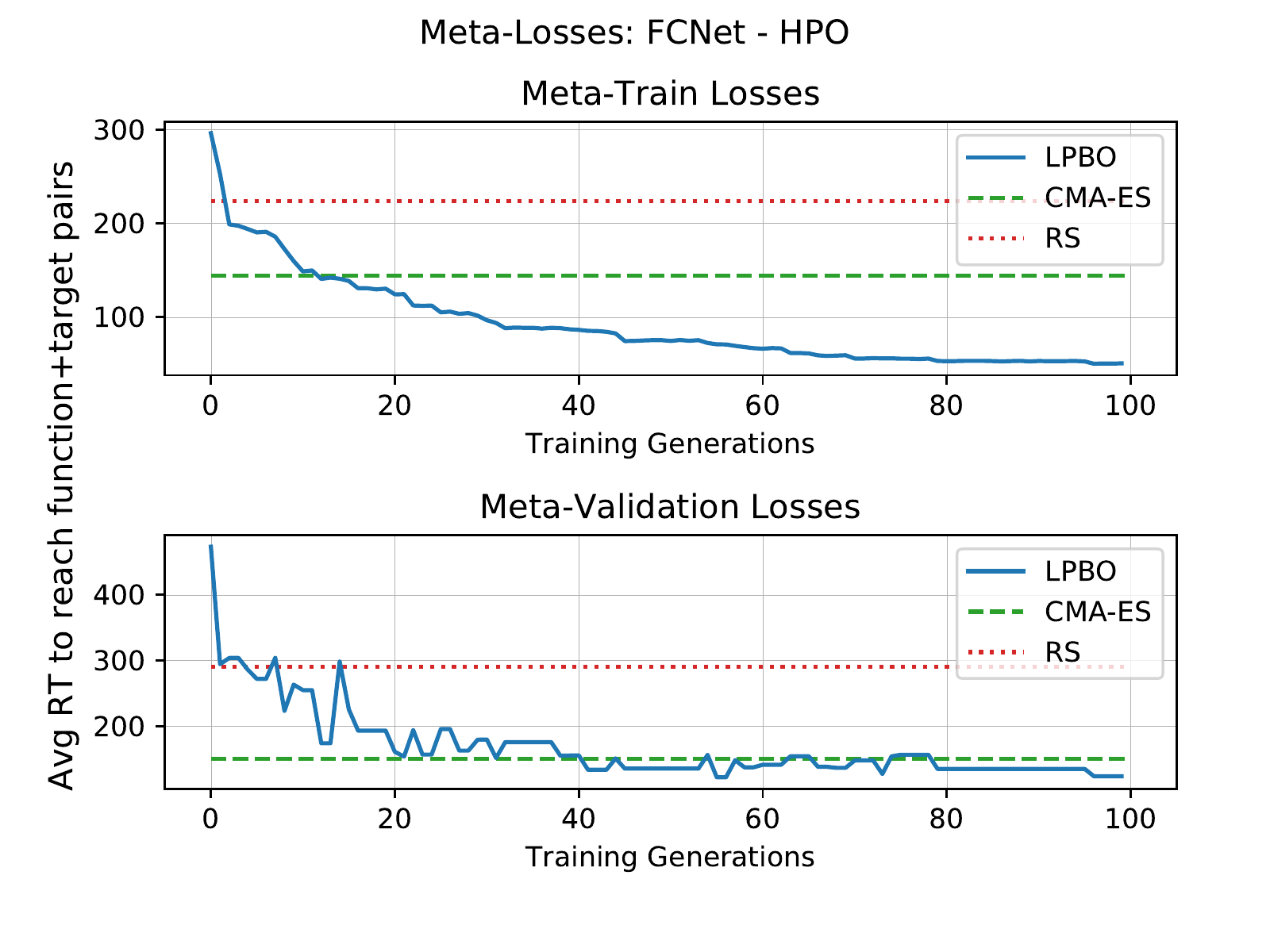}
\end{subfigure}\hfil 
\begin{subfigure}{0.3\textwidth}
  \includegraphics[width=\linewidth]{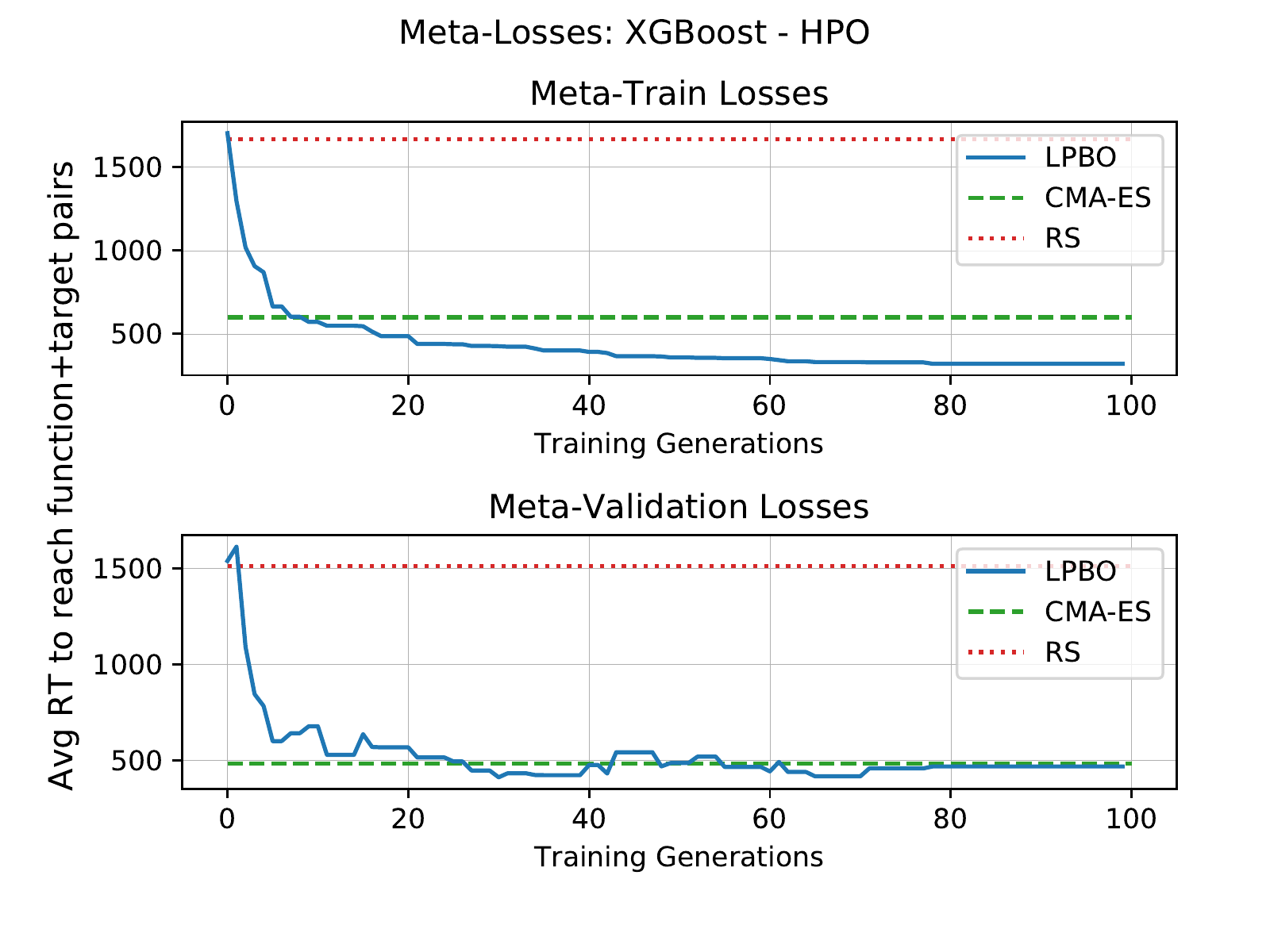}
\end{subfigure}
\caption{Results in meta-training HPO tasks. ECDFs (top) and meta-losses (bottom) for hyperparameter search of SVM (6D), a fully connected neural network (FC-Net, 6D) and XGBoost algorithm (8D).}
\label{fig:Experiments/HPO}
\end{figure*}

\subsection{Results Analysis}

In our experiments, three main elements on the proposed LTO-POMDP framework are discussed:

\begin{enumerate}
    \item \textbf{[Fit for purpose]} Is it able to train optimizers that perform well at solving the specific problems at hand?
    \item \textbf{[Generalization]} Is the learned optimizer capable of finding strategies to adapt to a variety of scenarios?
    \item \textbf{[Crossing the reality gap]} Can it be used for real-world optimization problems?
\end{enumerate}

For that purpose, three distinct scenarios are investigated. First, we train the optimizer using different instances of a given function (Sec.~\ref{sec:MIO}). This configuration is then extended to families of functions that share common properties for learning more general optimizers (Sec.~\ref{sec:MFO}). Finally, optimizers adapted to hyperparameter optimization for machine learning models are investigated as a real-word application (Sec.~\ref{sec:HPO}). All experiments compare the proposed LPBO with a random search, as a lower bound, and CMA-ES, which is known to be an effective well-performing generic BBO method.

\subsubsection{Scenario 1: Multiple Instances Optimization}\label{sec:MIO}

We explore the setting where the optimizer is learned from a set of instances of one function type on each experiment (see Fig.~\ref{fig:Experiments/BBOBSingle}). First results on Linear Slopes outlines that LPBO can identify the common global structure of related optimization tasks, with the optima being located at the boundaries of the domain. Both baselines (which do not have access to no such prior bias) fail to solve it, while LPBO outperforms all of the methods.

In the second setting, the learned optimizer outperforms the baselines when the problems require a good trade-off between local and global search. It reflects the awareness of the optimizer to a global structure where it can locate a promising region and result in an efficient local search strategy, typical of multi-modal functions with weak global structures such as the Schwefel and Lunacek bi-Rastrigin Functions. In comparison, our baselines tend to fall into a penalized area (e.g., the cliff in some Schwefell function regions) or fail to locate the promising regions.

In the last case, the Griewank-Rosenbrock function is used, which is a highly irregular multi-modal task. CMA-ES demonstrates notorious robustness to reach the optimal (i.e., it can solve all targets).  The inductive bias seems to limit the search for a precise local search of the optima due to the highly irregular regions of the function. However, the LPBO is outperforming the baselines to quickly find easier targets (i.e., the area under the curve of ECDF before 200 evaluations) being about three times faster. In practice, the user may decide whether it is acceptable to reach for a lower precision at an increased speed.

\subsubsection{Scenario 2: Multiple Functions Optimization}\label{sec:MFO}

Different meta-training scenarios are now explored for learning optimizers usable over a larger class of problems (see Fig.~\ref{fig:Experiments/BBOBGroup}). The two most challenging groups according to the COCO framework\footnote{https://coco.gforge.inria.fr/downloads/download16.00/bbobdocfunctions.pdf} are used: (1) multi-modal with adequate global structure functions and (2) multi-modal functions with weak global structure. We keep the same hyperparameters for the learned algorithm to show that zero (or minimal) fine-tuning is required for different black-box optimization tasks in real-world problems.

The comparative performance of the LPBO approach is significant in group 1. This scenario is particularly challenging for most algorithms due to these functions' deceptive and highly variable nature. The learned optimizer can no longer rely on simple strategies such as ``only sample points in the domain boundary'' or ``locate the plateau where the optima might be''. Therefore, the learned algorithm provides better optimization strategies with the right balance between local search and global awareness, leading to higher performances in dimensions 2, 5, and 10.

The second scenario (group 2) tests the method's ability to generalize over various global properties, where ``higher-level'' search strategies are needed. Interestingly, the method achieves good performances, only slightly outperformed by CMA-ES in dimension two but more sample efficient to solve easier targets. All three methods present similar results regarding their curve tendency and the final value in dimension 10, which may imply that further function evaluations are necessary for this scenario. 

\subsubsection{Hyperparameters Optimization}
\label{sec:HPO}

Finally, we explore our approach to the hyperparameters tuning of machine learning models (see Fig.~\ref{fig:Experiments/HPO}). The method outperforms the two baselines in three different scenarios of various models, corresponding to parameter search for a  Support Vector Machine (SVM, 2D), a fully connected neural network (FC-Net, 6D), and the extreme gradient boosting model (XGBoost, 8D). Examples of the SVM training landscape are shown in Fig.~\ref{fig:intro}. In higher dimensions, as in XGBoost, the LPBO is about twice as fast as than CMA-ES, which is often considered to have the most powerful self-adaptation mechanisms in BBO optimization \cite{last_one}. In all scenarios, the high performance of LPBO seems to outline the relevance of our method to design general search strategies applicable to diverse instances of machine learning algorithms.

\section{Conclusion}
\label{sec:6}

This paper proposes a data-driven approach for adaptable black-box population-based search. For that purpose, we introduce a novel framework for learning custom optimizers capable of efficiently solving optimization tasks. In particular, we establish a connection between black-box optimization methods and modeling of decision-making algorithms used in areas such as reinforcement learning, showing that learning an optimizer can be viewed as a policy search on a particular POMDP.

Most approaches proposed so far for learning to optimize are framed in gradient-based search (e.g., \cite{andrychowicz_learning_2016}, \cite{li_learning_2017}). In contrast, our solution is derivative-free and developed in population-based BBO scenarios. A key finding is our optimizer representation that does not use any information of the optimization tasks, except each evaluation ranking of the solutions. Therefore, it becomes invariant to some transformations of the functions. 

The learned optimizer relies on a meta-loss based on the performance of stochastic black-box algorithms, a novel neural architecture for the learned population-based optimizer, and a genetic algorithm as the meta-optimizer. Performances are evaluated on several scenarios: multiple instances of the same function, several functions with similar properties, and hyperparameter optimization of machine learning models. The learned optimizer is about two to three times faster than other approaches, and no fine-tuning of the method was needed in all experiments. While CMA-ES does not adapt quickly to several functions requiring hundreds of evaluations to reach roughly 50\% of the targets solved, the learned algorithm can reach 80\% of the targets in the Schwefel Function or even 100\% (in the Lunacek bi-Rastrigin function). This work is a promising avenue to automatically learn black-box optimizers for some specific contexts, achieving strong performance in solution quality and search efficiency. To conclude, we hope this work can encourage further development of learning optimization algorithms and augment the BBO toolbox in different scenarios.

\bibliographystyle{ACM-Reference-Format}
\bibliography{sample-bibliography} 

\end{document}